\documentclass[letterpaper]{article} % DO NOT CHANGE THIS
\usepackage{aaai25}  % DO NOT CHANGE THIS
\usepackage{times}  % DO NOT CHANGE THIS
\usepackage{helvet}  % DO NOT CHANGE THIS
\usepackage{courier}  % DO NOT CHANGE THIS
\usepackage[hyphens]{url}  % DO NOT CHANGE THIS
\usepackage{graphicx} % DO NOT CHANGE THIS
\usepackage{natbib}  % DO NOT CHANGE THIS AND DO NOT ADD ANY OPTIONS TO IT
\usepackage{caption} % DO NOT CHANGE THIS AND DO NOT ADD ANY OPTIONS TO IT
\usepackage{algorithm}
\usepackage{algorithmic}
\usepackage{amsmath}
\usepackage{amssymb}
\usepackage{booktabs}
\usepackage{multirow}
\usepackage{subfigure}
\usepackage{xcolor}

\newcommand{\methodname}{ISPF}

\newcommand{\prefilter}{PreFilter}
\newcommand{\postfilter}{PostFilter}
\newcommand{\inhibsyn}{Inhibited Synthesis}
\newcommand{\inhibsynShort}{IS}
\newcommand{\postfilterShort}{PF}

\def\chenhao{\textcolor{black}}

\usepackage{newfloat}
\usepackage{listings}
\DeclareCaptionStyle{ruled}{labelfont=normalfont,labelsep=colon,strut=off} % DO NOT CHANGE THIS
\floatstyle{ruled}
\newfloat{listing}{tb}{lst}{}
\floatname{listing}{Listing}
\title{Toward Efficient Data-Free Unlearning}
\author{
    Chenhao Zhang\textsuperscript{\rm 1},
    Shaofei Shen\textsuperscript{\rm 1},
    Weitong Chen\textsuperscript{\rm 2},
    Miao Xu\textsuperscript{\rm 1*}
}
\affiliations{
    \textsuperscript{\rm 1}University of Queensland,
    \textsuperscript{\rm 2}University of Adelaide\\
    \textsuperscript{\rm *}Corresponding Author\\
    
    chenhao.zhang@uq.edu.au, shaofei.shen@uq.edu.au, weitong.chen@adelaide.edu.au, miao.xu@uq.edu.au
}

\usepackage{bibentry}
\begin{document}

\maketitle

\begin{abstract}
Machine unlearning without access to real data distribution is challenging. The existing method based on data-free distillation achieved unlearning by filtering out synthetic samples containing forgetting information but struggled to distill the retaining-related knowledge efficiently. In this work, we analyze that such a problem is due to over-filtering, which reduces the synthesized retaining-related information. We propose a novel method, Inhibited Synthetic PostFilter (ISPF), to tackle this challenge from two perspectives: First, the Inhibited Synthetic, by reducing the synthesized forgetting information; Second, the PostFilter, by fully utilizing the retaining-related information in synthesized samples. Experimental results demonstrate that the proposed ISPF effectively tackles the challenge and outperforms existing methods.
\end{abstract}

\begin{links}
    \link{Code}{https://github.com/ChildEden/ISPF}
    % \link{Datasets}{https://aaai.org/example/datasets}
    \link{Extended version}{https://arxiv.org/abs/0000.00000}
\end{links}

\section{Introduction}

Machine unlearning ~\cite{SISA,unlearningSurvey1,unlearningSurvey2} is an emerging paradigm that enables machine learning models to selectively forget training data, primarily to enhance user privacy and comply with regulations~\cite{righttobeforgotten,california}, or to correct errors within the dataset~\cite{advApp,advApp1}. Existing unlearning methods typically require access to the original training data to accurately discern which information needs to be removed (the ``forgetting'' data) and which must be preserved (the ``retaining'' data). This access is crucial for precisely identifying the parameters impacted by the forgetting data, ensuring that only these parameters are adjusted~\cite{SalUN,SSD} while the functionality of the model is maintained for the retaining data~\cite{unrolling,SCRUB}. However, in practice, the original training data may not always be available due to reasons such as compliance with privacy laws that mandate data deletion, or organizational policies aimed at optimizing storage efficiency. Moreover, alternative data sources that share a similar distribution with the training data might also be inaccessible due to legal restrictions. Despite these challenges in data access, requests for unlearning tasks such as forgetting specific concepts or classes of data can still be initiated; this scenario, where unlearning is required without access to the original or similar training data, is termed \emph{data-free unlearning}.

An existing method addressing data-free unlearning is Generative Knowledge Transfer (GKT)~\cite{GKT}, which utilizes Data-free Knowledge Distillation~\cite{ZSKD} to selectively transfer knowledge from a trained model to an unlearned model. Both data-free unlearning and data-free knowledge distillation (DFKD) train a generator with the principle of synthesizing new samples that simulate the original training distribution, which can be achieved by harnessing the capabilities of the well-trained model, and followed by the knowledge distillation~\cite{DAFL,DFQ,adv-DFKD}. 
With meaningful synthetic samples in hand, GKT selectively distil only the necessary retaining knowledge by filtering out synthetic samples potentially belonging to the forgetting class. Specifically, it excludes samples where the probability associated with forgetting classes as outputted by the logits exceeds a very low threshold before distillation. This filtering strategy can ensure that samples involved in distillation contain minimal forgetting class information and effectively filter out forgetting knowledge. 

\begin{figure}[t]
\centering
    \subfigure[Without filter]{
        \begin{minipage}[t]{0.45\linewidth}
            \centering
            \includegraphics[width=\linewidth]{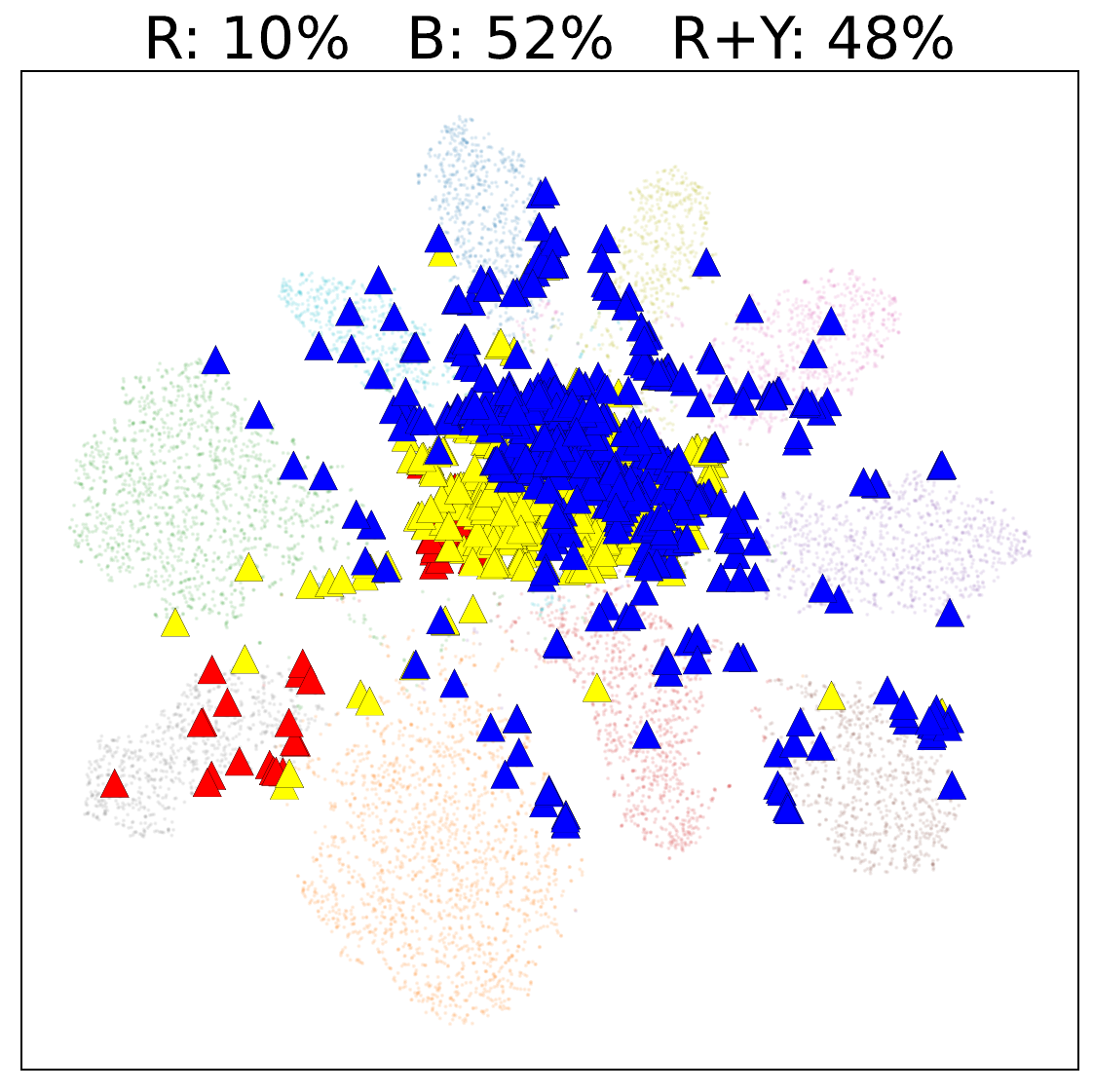}
        \end{minipage}
        \label{fig:problem_DFKD}
    }
    \hfill
    \subfigure[With filter]{
        \begin{minipage}[t]{0.45\linewidth}
            \centering
            \includegraphics[width=\linewidth]{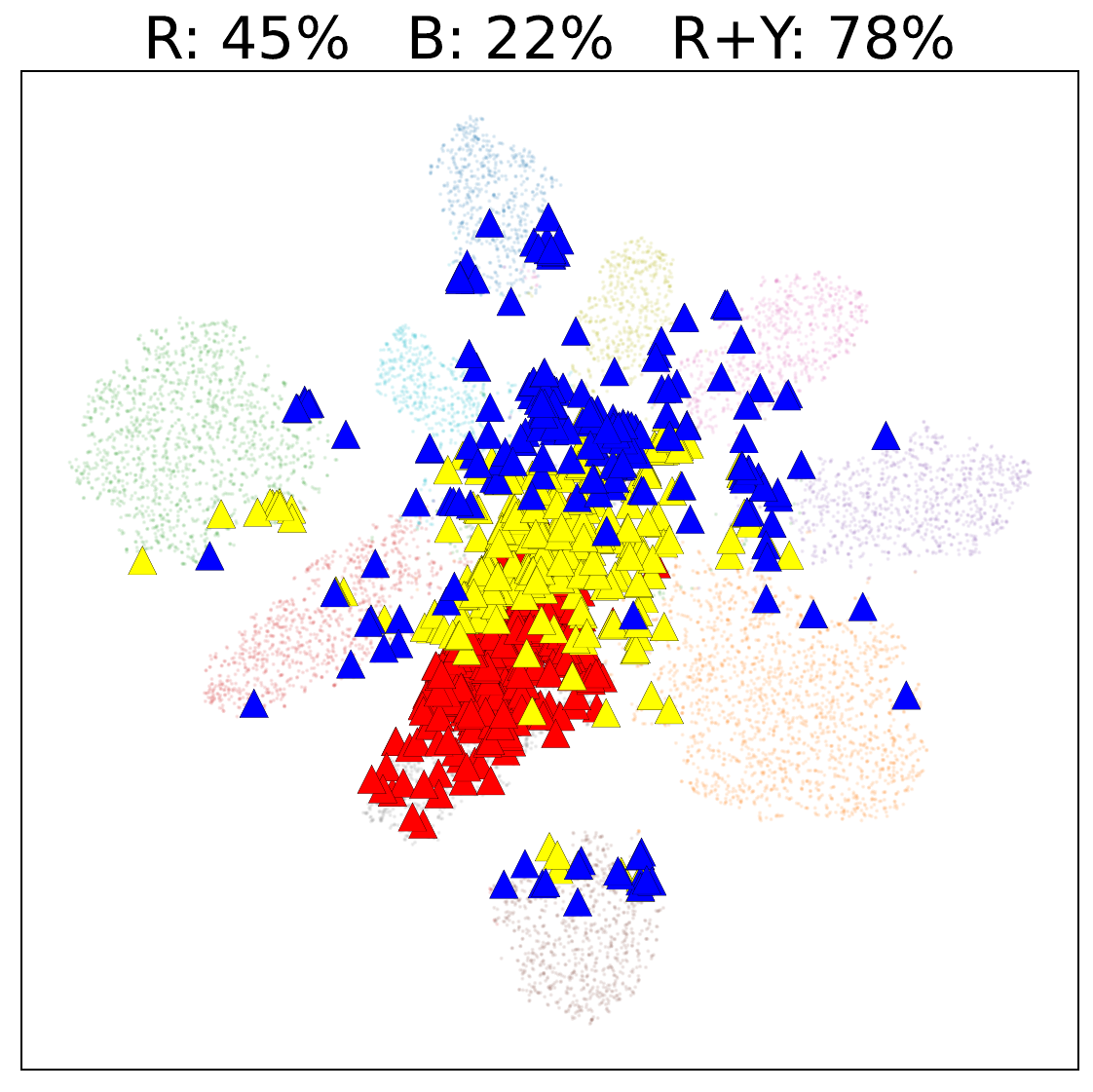}
        \end{minipage}
        \label{fig:problem_GKT}
    }
    \caption{
        Comparative Visualization of Synthetic Samples during the DFKD process on the SVHN Dataset:
        \textbf{Background shadow} illustrates the real data distribution;       \textbf{Triangles:} Synthetic samples; \textbf{Red (R):} Samples of the forgetting class (digit ``7''); \textbf{Yellow (Y):} Non-forgetting class samples but still filtered out by the filter of GKT; \textbf{Blue (B):} Samples deemed suitable for participation in the distillation process. 
        \textbf{(a)} When there is no filter, i.e., the distillation of complete knowledge, the synthesized data has minimal bias toward the forgetting class.
        \textbf{(b)} When a filter is used, i.e. when performing unlearning, a biased high volume of the forgetting class sample is synthesized.
    }
\label{fig:problem}
\end{figure}

Despite selective distillation being addressed, the training principle of DFKD's generator, which with the aim of synthesizing ``not yet seen" data for student networks, can present a novel challenge in unlearning scenarios. Intuitively, in the context of unlearning, forgetting class samples will inevitably be the ``not yet seen" data for the student network, resulting the generator increasingly producing a significant number of samples from the forgetting class. As shown in Figure~\ref{fig:problem}, when the filtering strategy is employed, the generator can synthesize an increasing number of forgetting class samples. Consequently, the large volume of filtered-out potential forgetting class samples significantly reduces the pool of data available for distillation, leading to inefficiencies. Furthermore, this filtering impedes the distillation process by inadvertently excluding logits outputs from other classes, which are crucial for knowledge transfer and represent a key advantage of knowledge distillation that relies on these outputs instead of hard labels.

In this work, we first strictly analyze the efficiency problem when using Data-free Knowledge Distillation (DFKD) to perform selective distillation and achieve data-free unlearning. We find that enriching the information about retaining classes involved in the distillation process can effectively improve student's learning of retaining-related knowledge. We propose the Inhibited Synthetic PostFilter (\methodname) method to achieve this goal from two perspectives. The first is to reduce the synthetic potential forgetting class sample, which is named \inhibsyn~(\inhibsynShort), by designing a new objective function for training the generator. This objective can effectively suppress the generator's exploration of the forgetting classes' distribution. The second is to involve all synthetic samples in the distillation process and filter out forgetting class knowledge by modifying the teacher's output, which is named \postfilter~(\postfilterShort). This allows the distillation process to leverage as much information as possible about the retaining classes in a synthesis batch. We summarize our contribution as follows:
\begin{itemize}
    \item We identify and analyze the challenge associated with applying DFKD in unlearning. Our findings suggest that by enriching the information related to retaining classes in the distillation process, we can significantly improve the student model's acquisition of knowledge pertaining to these classes.
    \item We propose two key technologies for tackling this challenge: 1) \inhibsyn~for reducing the synthesis of information about the forgetting class and 2) \postfilter~for leveraging as much information as possible about the retaining classes.
    \item Experimental results show that our proposed method can tackle the challenge and outperform existing methods.
\end{itemize}

\chenhao{Due to space constraints, the related works section is provided in Appendix~\ref{sec:extendRelatedWorks} of the extended version.}

\section{Preliminaries}
Given a dataset $\mathcal{D}=\{(x_i,y_i)|x_i\in\mathcal{X}, y_i\in\mathcal{Y}\}_i^N$, where the data instance $x\in\mathbb{R}^d$ is sampled from a real distribution $P(x)$, i.e., $x\sim P(x)$. A neural network $h(\theta,x)$ which is parameterized by $\theta$ can output a probability vector that indicates the probability of the given $x$ being classified as each class, i.e., $h(\theta,x)\in (0,1)^K$, where $K$ is the number of classes. When the network is required to unlearn classes $\mathcal{Y}_f$, the classes that need to be retained are in $\mathcal{Y}_r$, where $\mathcal{Y}_f\cup\mathcal{Y}_r=\mathcal{Y}$ and $\mathcal{Y}_f\cap\mathcal{Y}_r=\emptyset$. We can further define the corresponding subsets $\mathcal{D}_f$ and $\mathcal{D}_r$, where $\mathcal{D}_f=\{(x,y)|y\in\mathcal{Y}_f\}$ and $\mathcal{D}_r=\{(x,y)|y\in\mathcal{Y}_r\}$.
Most existing unlearning methods, which we denote as $\mathcal{U}$, require data samples from the real distribution, and their goal can be formed as 
$\theta_{un}=\mathcal{U}(\theta_{or},x)$, where 
$\theta_{or}$ is the parameter of the original model.

\subsubsection{Data-free Unlearinng}
A data-free unlearning method is required to take only the trained network $h(\theta_{or},\cdot)$ without sample $x\sim P(x)$, i.e., $\theta_{un}=\mathcal{U}(\theta_{or})$.

\subsubsection{Data-free Knowledge Distillation (DFKD)}
Knowledge Distillation (KD) can transfer knowledge from a well-trained teacher network $h(\theta_T,\cdot)$ to a randomly initialized student network $h(\theta_S,\cdot)$. For representation convenience, we denote them as $T(\cdot)$ and $S(\cdot)$, respectively. The general goal of KD is to minimize the gap between the outputs of these two networks by using the Kullback–Leibler divergence~\cite{KLdiv} to measure the discrepancy. In DFKD, the input samples $\widetilde{x}$ are synthesized by a generator, thus the student's loss function is
\begin{equation}
\label{eq:KD}
    \mathcal{L}_S(\widetilde{x})=D_{KL}(T(\widetilde{x})\parallel S(\widetilde{x})),
\end{equation}
where $\widetilde{x}=G(z)$ and the $G(z)$ is a generator that
can project a given lower-dimensional noise $z\in N(0,I)^{d_z}$ to $\widetilde{x}\in\mathbb{R}^d$. 
DFKD methods usually train the generator adversarially by maximizing the output distribution discrepancy between teacher and student, i.e.,
\begin{equation}
\label{eq:advloss}
    \mathcal{L}_{adv}(\widetilde{x})=-D_{KL}(T(\widetilde{x})\parallel S(\widetilde{x})).
\end{equation}

\subsubsection{\prefilter}
In addition to using the same objectives as Eq.~\ref{eq:advloss} to train the generator, the existing data-free unlearning method, GKT~\cite{GKT}, applies a filter ahead of the generator which receives all the synthetic samples and filters them out before passing them to the student and we refer to this as \prefilter. Specifically, the synthetic samples that can be involved in the distillation satisfy
\begin{equation}
    \forall \widetilde{x}_i,\forall k\in\mathcal{Y}_f, T_k(\widetilde{x}_i)<\delta,
\end{equation}
where $k$ is the class label, $T_k(\cdot)$ is the $k$-th value of the output probability vector and $\delta$ is a hyperparameter whose exact value for ten-class datasets in the GKT is 0.01.

\section{Method}
This section is structured as follows: Section~\ref{sec:anayzeAdv} analyzes the challenge encountered in unlearning scenarios using DFKD directly. Section~\ref{sec:IS} and Section~\ref{sec:PF} introduce the two key techniques, i.e., \inhibsyn~and \postfilter, respectively. The overall algorithm is placed in Appendix~\ref{sec:algo}.

\subsection{Challenge in Unlearning}
\label{sec:anayzeAdv}
We first expand the form of Eq.~\ref{eq:advloss} into the following form
\begin{equation}
\label{eq:expandAdvLoss}
    \mathcal{L}_{adv}(\widetilde{x}_i)=-\sum_k T_k(\widetilde{x}_i)\cdot[\log(T_k(\widetilde{x}_i))-\log(S_k(\widetilde{x}_i))],
\end{equation}
where $\widetilde{x_i}=G(z_i)$. 
Note that this is an objective function for updating $G$, thus, the $T$ and $S$ are fixed when computing $\mathcal{L}_{adv}$.
The goal of minimizing the Eq.~\ref{eq:expandAdvLoss} is equal to maximizing the $-\mathcal{L}_{adv}$
\begin{equation}
\label{eq:expandAdvLoss1}
\begin{split}
    & \min_{G}\mathbb{E}_{\widetilde{x}=G(z)}\left[\mathcal{L}_{adv}(\widetilde{x})\right] \\
    =&\max_{G}\mathbb{E}_{\widetilde{x}=G(z)}\left[-\mathcal{L}_{adv}(\widetilde{x})\right] \\
    =&\max_{G}\mathbb{E}_{\widetilde{x}=G(z)}\left[\sum_k T_k(\widetilde{x})\cdot\left[\log(T_k(\widetilde{x}))-\log(S_k(\widetilde{x}))\right]\right]
\end{split}
\end{equation}

The expected result for the student is that there is no high enough output probability of forgetting class for any data sample. Therefore, for any forgetting class $f$, the $f$-th element of the student's output is very low
\begin{equation}
    \forall \widetilde{x}=G(z),\; 0<S_f(\widetilde{x})<\epsilon.
\end{equation}
To reach the goal of maximizing Eq.~\ref{eq:expandAdvLoss1} when $S_f(\widetilde{x})$ is small it is necessary to increase $T_f(\widetilde{x})$ to a large value.
This leads to an increasing probability that synthetic samples will determined by the teacher as forgetting class.
As the training processes, an increasing number of synthetic data $\widetilde{x}$ will be filtered out due to the $T_f(\widetilde{x})$ exceeding the specified threshold $\delta$ of the \prefilter, resulting in a decreasing number of samples being used for distillation. This ultimately leads to a reduction in distillation efficiency.

\subsection{\inhibsyn~(\inhibsynShort)}
\label{sec:IS}
As analyzed above, the student's lack of familiarity with the synthetic samples, which contain information regarding the forgetting class, encourages $G$ to explore further into samples that contain a greater quantity of information regarding the forgetting class. This results in a greater number of synthetic samples being filtered out by \prefilter~before distillation. Therefore, we want the generator to synthesize fewer samples of the forgetting class and more samples of the retaining class. To address this issue, we need to encourage the generator to explore samples that are not previously encountered by the student, while simultaneously suppressing the generator's exploration of the underlying real distribution of the forgetting class. 

To minimize the number of samples generated by $G$ that contain information about the forgetting class, it is necessary to reduce the value of $T_f(\widetilde{x})$. We can achieve this by reducing the gap between $T_f$ and $S_f$ given that $S_f(\widetilde{x})$ is always low for each synthetic sample. Therefore, we propose the inhibited synthesis loss for the generator's learning, i.e.,
\begin{equation}
    \begin{split}
    \label{eq:IA}
        \mathcal{L}_{IS}(\widetilde{x}_i) = &-\sum_{k\in\mathcal{Y}_r} T_k(\widetilde{x}_i)\cdot[\log(T_k(\widetilde{x}_i))-\log(S_k(\widetilde{x}_i))] \\
        &+\sum_{f\in\mathcal{Y}_f}  T_f(\widetilde{x}_i)\cdot[\log(T_f(\widetilde{x}_i))-\log(S_f(\widetilde{x}_i))].
    \end{split}
\end{equation}
Given that samples with $T_f(\widetilde{x})$ exceeding a very small threshold $\epsilon$ are excluded from distillation, leaving only $\widetilde{x}$'s with $0<T_f(\widetilde{x})<\epsilon$, therefore, the $S_f(\widetilde{x})$ of a student engaged in learning with the objective of Eq.~\ref{eq:KD} will invariably be smaller than $\epsilon$. Consequently, we can use the Eq.~\ref{eq:IA} to actively diminish the $T_f(\widetilde{x})$, thereby impeding the generator from producing samples with elevated $T_f(\widetilde{x})$.

\subsection{\postfilter~(\postfilterShort)}
\label{sec:PF}
In the previous discussion, we endeavored to diminish the number of samples excluded from distillation by impeding $G$'s synthesis of samples bearing information about the forgetting class, thereby augmenting the number of samples utilized for distillation. The experiments revealed that even when the inhibited $G$ has synthesized almost no samples that can be classified as the forgetting class by the teacher, some samples are still filtered out (third column in Figure~\ref{fig:efficiency}). 
This is because the synthetic samples lack sufficient purity to represent an individual class. Synthetic samples belonging to the retaining class may contain information about the forgetting class, and similarly, synthetic samples belonging to the forgetting class may also contain information about retaining classes. In light of this observation, we wished to enrich the material employed in the distillation process by fully using the synthetic samples.

Specifically, we construct new supervision information for the student by redistributing the logits of the teacher output. This is achieved by setting the forgetting classes' value in the teacher's output logits to the lowest value and distributing the sum of the subtracted logits evenly to the logits of the retaining classes. We denote the logits output (network's output before Softmax) of each sample from the teacher as $\boldsymbol{t}$. We first calculate the total logits value $\Delta$ that needs to be redistributed by summing the difference between forgetting class logits and the minimal value in $\boldsymbol{t}$, i.e., 
\begin{equation}
    \Delta=\sum_{k\in \mathcal{Y}_f}\left[\boldsymbol{t}_k - min(\boldsymbol{t})\right]
\end{equation}
Then, we construct the supervision information $\hat{\boldsymbol{t}}$ by redistributing the $\Delta$ to all retaining classes and setting the forgetting class logits value as the minimum, i.e.,
\begin{equation}
     \hat{\boldsymbol{t}}_{k}=
    \begin{cases}
        \boldsymbol{t}_{k}+\frac{\Delta}{K-|\mathcal{Y}_f|}, &\quad\text{if }k\notin\mathcal{Y}_f, \\
        min(\boldsymbol{t}), &\quad\text{otherwise}.
    \end{cases}
\end{equation}
The student will then use the redistributed teacher logits as supervision information to calculate losses
\begin{equation}
\label{eq:KD_ReD}
    \mathcal{L}_S(\widetilde{x})=D_{KL}(Softmax(\hat{\boldsymbol{t}})\parallel S(\widetilde{x})).
\end{equation}

A simpler implementation can be setting the value of $T(\widetilde{x})$ corresponding to the forgetting class as 0 and renormalizing other values to obtain the distillation target. We experimentally compared this simple implementation with our proposed \postfilterShort~(Appendix~\ref{sec:replacePostF}), and the results show that \postfilterShort~outperforms this simple implementation in terms of both unlearned model's performance and unlearning guarantee.

\section{Experiment}
We evaluate the effectiveness of the \methodname~which is composed of two proposed techniques, the \inhibsyn~(\inhibsynShort) and \postfilter~(\postfilterShort), on three widely used benchmark datasets, i.e., SVHN~\cite{svhn}, CIFAR-10 and CIFAR-100~\cite{cifar10}. Two neural network architectures, i.e., AllCNN~\cite{allcnn} and ResNet18~\cite{resnet}, are used in our experiments. For SVHN and CIFAR-10 datasets, we apply both network architectures. For CIFAR-100, we apply only ResNet18. Implementation details are in the Appendix~\ref{sec:hyperparam}. 

\subsection{Experimental Setup}
\subsubsection{Datasets}
We perform unlearning on each class from both SVHN and CIFAR-10. CIFAR-100 has 100 classes which can be divided into 20 super-classes. We perform single-class unlearning on three randomly selected classes of the CIFAR-100, i.e., $y_f\in\{18,33,79\}$. We also use the CIFAR-100 for multi-classes unlearning experiments where more than one class needs to be unlearned. We use all 100 classes to train the original model for generality and choose one from each of 10 different super-classes to make up the 10 classes to unlearn. Specifically, we select class labels $\mathcal{Y}_f=\{0,1,2,3,4,5,6,8,9,12\}$, and a more detailed explanation about why selecting these classes is presented in the section of \textbf{CIFAR-100 Results}.

\subsubsection{Baselines}
We refer to the model retrained from scratch as the gold result and compare the proposed methods with the existing data-free unlearning method that requires no access to data from real distribution, the \textbf{GKT}~\cite{GKT}. The naive method of blocking synthetic samples that can be determined as forgetting class by the original model has also been included in baselines as \textbf{BlockF}.

\subsubsection{Fundamental DFKD Method}
We select the DFQ~\cite{DFQ} as the main fundamental DFKD method because it is a representative DFKD method that obeys the adversarial inversion-and-distillation paradigm and has been widely used as a baseline in other DFKD works. In the following experimental sections, DFKD refers to DFQ unless otherwise stated. We also conducted experiments using ZSKT~\cite{ZSKD}, which was originally used by the GKT, as the fundamental DFKD (Appendix~\ref{sec:zsktRes}). 

\subsubsection{Evaluation Metrics}
To evaluate the effectiveness of our proposed methods, we refer to existing works~\cite{GKT,boundaryUnlearning,UNSIR,nullSpace,SalUN} and select the following four metrics.

\textit{Classification Accuracies:} We test unlearned models on the forgetting test data $\mathcal{D}_f^{test}$ and the retaining test data $\mathcal{D}_r^{test}$ for obtaining accuracies $A_f$ and $A_r$, respectively. Both $A_f$ and $A_r$ are the closer to that of the \textit{Retrain} model the better.

\textit{Anamnesis Index (AIN):} As introduced by~\citet{GKT}, the AIN evaluates how much forgetting information remains in an unlearned model by measuring the amount of time (training steps) it takes to relearn a comparable $A_f$ as the original model. This is a ratio of an unlearned model's relearning time to a retrained model's relearning time. Therefore, the closer the ratio is to 1, the better. 

\textit{Membership Inference Attack (MIA):} We implement two types of MIA attack. 1) Following~\cite{SalUN,nullSpace}, we define the first MIA metric MIA$_{\uppercase\expandafter{\romannumeral1}}$ as the rate of unlearning samples that are identified as not being in the training set of the unlearned model. A higher MIA$_{\uppercase\expandafter{\romannumeral1}}$ indicates a more effective unlearning. 2) We also refer to experiments in Boundary Unlearning~\cite{boundaryUnlearning} and implement a simple general MIA based on~\cite{mia}. 
In this approach, multiple shadow models are trained to gather signals for the attacker's training. The attacker aims to determine whether a given signal originates from a sample within the in-training dataset. During the attacker's testing, a test set that includes both in-training and out-of-training data is used and we define the MIA$_{II}$ as the F1 score when it attacks each unlearned model.
Generally, the attacker can achieve the best performance on the signals obtained from the original model, however, a lower MIA$_{II}$ doesn't mean it's a better unlearning guarantee, as a randomly initialized model provides signals that can confuse the attacker. Therefore MIA$_{II}$ should be close to the F1 when attacking the retrained model.

\begin{table*}[t]
\centering
\setlength\tabcolsep{4pt}
\begin{tabular}{c|c|ccccc|ccccc}
\toprule
\multirow{2}{*}{Arch}  & \multirow{2}{*}{Method} & \multicolumn{5}{|c|}{SVHN}    & \multicolumn{5}{c}{CIFAR-10} \\
                          &                         & $A_f$ & $A_r$ & MIA$_{I}$ & MIA$_{II}$ & AIN & $A_f$  & $A_r$ & MIA$_{I}$ & MIA$_{II}$ & AIN \\
\midrule
\multirow{8}{*}{\rotatebox{90}{AllCNN}} & Original  & 93.86$\pm$0.27  & 93.86$\pm$0.27            & 4.96    & 39.67             & -                 & 90.91$\pm$0.13  & 90.91$\pm$0.13            & 2.99          & 45.86             & -   \\
                          & DFKD                    & 91.37$\pm$0.23  & 91.37$\pm$0.23            & 0.0     & 39.56             & 0.07              & 84.85$\pm$0.15  & 84.85$\pm$0.15            & 1.2           & 44.91             & 0.11   \\ \cline{2-12}
                          & Retrain                 & 0.0$\pm$0.0     & 94.14$\pm$0.13            & 100.0   & 33.28             & 1.0                 & 0.0$\pm$0.0     & 91.39$\pm$0.04            & 100.0         & 37.34             & 1.0      \\
                          & BlockF                  & 90.33$\pm$0.71  & 92.35$\pm$0.07            & 1.91       & 38.75             & 0.06              & 65.67$\pm$1.45  & 85.08$\pm$0.5             & 10.4          & 44.65             & 0.10   \\
                          & GKT                     & 0.0$\pm$0.0     & 55.23$\pm$2.42            & 75.0    & 19.86             & 1.30              & 0.0$\pm$0.0     & 57.62$\pm$1.13            & 47.42         & 21.51             & 1.44   \\
              & \inhibsynShort~(ours)               & 0.0$\pm$0.0     & 90.88$\pm$0.21            & \textbf{100.0}   & \textbf{33.51}    & \textbf{0.99}     & 0.0$\pm$0.0     & \underline{84.93$\pm$0.02}& \textbf{100.0}& \textbf{37.35}    & \underline{0.86}   \\
                    & \postfilterShort~(ours)       & 0.0$\pm$0.0     & \underline{91.15$\pm$0.85}& \textbf{100.0}   & \underline{34.24} & 0.91              & 0.0$\pm$0.0     & 76.76$\pm$0.06            & \textbf{100.0}& 26.36             & 1.28   \\
        & \methodname~(ours)    & 0.0$\pm$0.0     & \textbf{92.68$\pm$0.13}   & \textbf{100.0}   & 34.54             & \underline{0.96}  & 0.0$\pm$0.0     & \textbf{86.02$\pm$0.03}   & \textbf{100.0}& \underline{36.14} & \textbf{0.89}   \\
\midrule
\multirow{8}{*}{\rotatebox{90}{ResNet18}} & Original& 94.34$\pm$0.11  & 94.34$\pm$0.11            & 2.55    & 44.59              & -                & 92.02$\pm$0.51  & 92.02$\pm$0.51            & 2.94          & 42.17             & -   \\
                          & DFKD                    & 91.72$\pm$0.38  & 91.72$\pm$0.38            & 0.83    & 38.34              & 0.17             & 83.25$\pm$0.59  & 83.25$\pm$0.59            & 0.0           & 39.67             & 0.04   \\ \cline{2-12}
                          & Retrain                 & 0.0$\pm$0.0     & 94.43$\pm$0.03            & 100.0   & 36.51              & 1.0                & 0.0$\pm$0.0     & 92.4$\pm$0.23             & 100.0         & 30.81             & 1.0   \\
                          & BlockF                  & 74.97$\pm$1.55  & 91.87$\pm$0.03            & 51.35   & 36.51              & 0.12             & 48.5$\pm$0.84   & \underline{81.49$\pm$0.44}& 6.47          & 36.21             & 0.04   \\
                          & GKT                     & 0.0$\pm$0.0     & 88.75$\pm$1.43            & \textbf{100.0}   & 29.07              & \underline{0.48} & 0.0$\pm$0.0     & 58.11$\pm$0.76            & 45.0          & 16.25             & \textbf{0.51}   \\
              & \inhibsynShort~(ours)               & 0.0$\pm$0.0     & 90.75$\pm$0.55            & \textbf{100.0}   & \textbf{34.27}     & 0.46             & 0.0$\pm$0.0     & 80.54$\pm$0.25            & \textbf{100.0}& \underline{28.11} & 0.22   \\
                    & \postfilterShort~(ours)       & 0.0$\pm$0.0     & \underline{91.61$\pm$0.41}& \textbf{100.0}   & 32.28              & \textbf{0.52}    & 0.0$\pm$0.0     & 81.21$\pm$0.72            & \textbf{100.0}& 22.73             & \underline{0.45}   \\
        & \methodname~(ours)    & 0.0$\pm$0.0     & \textbf{91.92$\pm$0.23}   & \textbf{100.0}   & \underline{33.67}  & 0.45             & 0.0$\pm$0.0     & \textbf{83.33$\pm$0.81}   & \textbf{100.0}& \textbf{29.36}    & 0.28   \\
\bottomrule
\end{tabular}
\caption{Unlearning performance averaged across all classes. The AIN is reported as a ratio and all other metrics are reported as percentages (\%). The \textbf{bold} record indicates the best result and the \underline{underlined} record indicates the second best result.}
\label{tab:mainRes}
\end{table*}

\subsection{Comparison with baselines}

\subsubsection{Unlearned Model Performance}
\begin{figure*}[t]
    \subfigure[SVHN-AllCNN]{
        \begin{minipage}[t]{\linewidth}
            \centering
            \includegraphics[width=\linewidth]{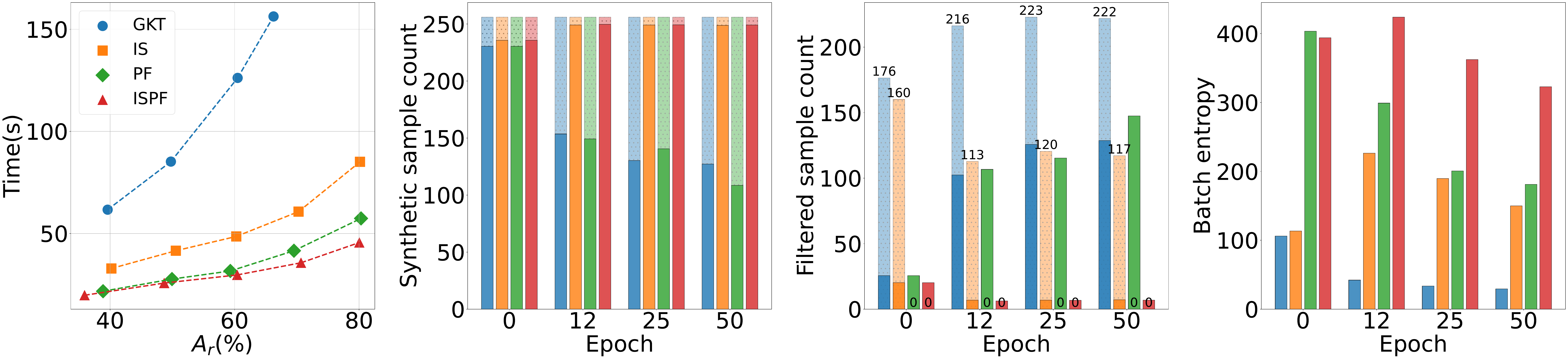}
        \end{minipage}
        \label{fig:svhnResFig}
    }
    \subfigure[CIFAR10-AllCNN]{
        \begin{minipage}[t]{\linewidth}
            \centering
            \includegraphics[width=\linewidth]{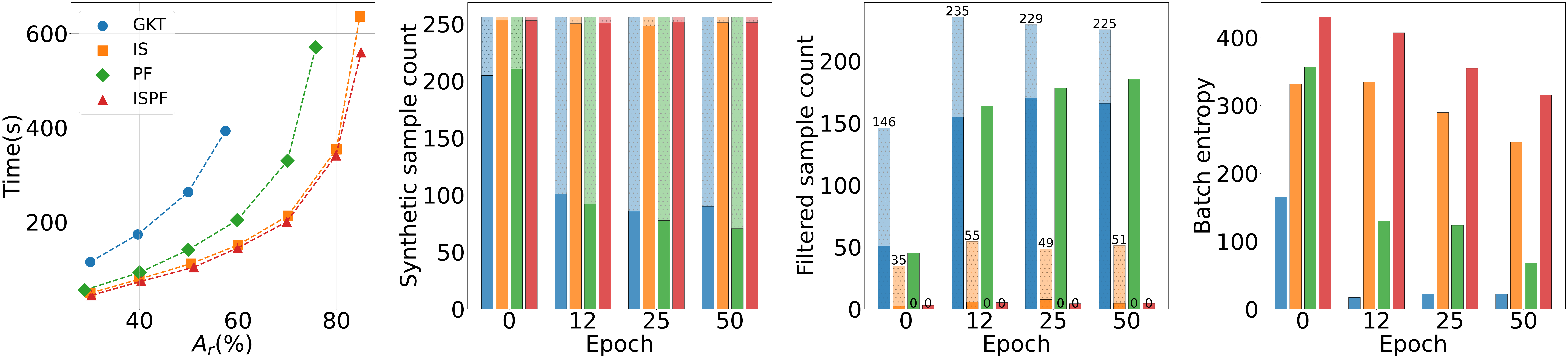}
        \end{minipage}
        \label{fig:cifar10ResFig}
    }

\caption{The colors in all figures are used to distinguish the different methods. The top row shows the results under the SVHN-AllCNN setting, and the bottom row shows the results under CIFAR10-AllCNN. The \textbf{first column} shows the results for $A_r$ vs. wall time. In the \textbf{second column}, the light bar filled with dots shows the number of synthetic samples classified as forgetting classes by the original model, and the dark bar shows the number of retaining class samples. In the \textbf{third column}, the light bar filled with dots shows the number of samples filtered out before distillation and the number indicates the exact number of filtered-out samples, and the darker bar shows the number of synthetic forgetting class samples, which is the same as the lighter bar in the second column. Corresponding results on ResNet18 are in the Appendix~\ref{sec:resnet18Res}.}
\label{fig:efficiency}
\end{figure*}

Table~\ref{tab:mainRes} reports unlearning performance averaged across all classes, and we also provide detailed results when each class is set as the forgetting class in Appendix~\ref{sec:detailRes}. Table~\ref{tab:mainRes} illustrates that the unlearned model obtained by BlockF still exhibits generalization ability to the forgetting class. This suggests that solely blocking samples, which are identified as the forgetting class by the original model, from participating in the distillation process can not effectively way block the forgetting class information. This is because the synthetic samples cannot represent an individual class in a sufficiently distinct manner, and the student can still learn the forgetting class through the samples that are determined as other classes. The GKT uses \prefilter~to filter samples based on the teacher's confidence in the forgetting class. This approach ensures that only samples with minimal confidence in the forgetting class are included in the distillation process, effectively filtering out knowledge related to the forgetting class in comparison to BlockF. This can be seen by comparing the $A_f$ performance of GKT and BlockF. However, the use of the \prefilter~will cause the increase in synthesizing forgetting class samples and also the exclusion of samples from other classes (third column in Figure~\ref{fig:efficiency}), which in turn leads to a reduction in the knowledge related to the retaining class participating in the distillation and affects the learning efficiency of retaining knowledge. For example, the $A_r$ of GKT is significantly inferior to that of DFKD which strives to facilitate the complete transfer of knowledge from the teacher to the student. 

In contrast, our proposed \methodname~outperformances baselines on both $A_r$ and $A_f$. This is because \methodname~effectively inhibits the synthesis of forgetting class samples through \inhibsynShort, while using \postfilterShort~to fully utilize the retaining-related knowledge in the synthetic samples. We will further explain why the \inhibsynShort~and \postfilterShort~work in the \textbf{Ablation} section.

\subsubsection{Unlearning Guarantee}
We note that in some settings, a single metric may not effectively differentiate the discrepancy in unlearning guarantee between methods. We therefore include three widely used metrics for evaluating the unlearning guarantee of unlearned models. Our proposed methods consistently outperform the comparative method on two or even three metrics across diverse settings. In Appendix~\ref{sec:unlearnedModelRes}, we present additional perspectives, including the model's predictive distribution and representation, to demonstrate that our proposed method exhibits performances more closely aligned with \textit{Retrain}.

\textbf{1) MIA$_{I}$: }
According to~\cite{SalUN,nullSpace}, the MIA$_{I}$ metric is defined as the rate of unlearning samples that are identified as not being in the training set of the unlearned model. As shown in Table~\ref{tab:mainRes}, the MIA$_{I}$ of the retrain model is 100\%, indicating that the attacker believes that none of the forgetting samples are included in the retrain model's training data. This result can also be observed in the Figure 2 in~\cite{nullSpace}. Conversely, models with all classes' knowledge exhibit comparable and relatively low performance on MIA$_{I}$. For instance, the MIA$_{I}$ of the original model and the model obtained through DFKD, both fall below 5\%. In comparison methods, the BlockF exhibits an $A_f$ that is not low, and MIA$_{I}$ is comparable to the original model and the one obtained by complete distillation. GKT's MIA$_{I}$ is considerably higher than BlockF's but remains significantly lower than Retrain's. Our proposed \methodname's MIA$_{I}$ is consistent with Retrain's, indicating that \methodname~has a superior unlearning guarantee in terms of MIA$_{I}$.

\textbf{2) MIA$_{II}$: }
From Table~\ref{tab:mainRes}, models with comprehensive knowledge of all classes exhibit the highest MIA$_{II}$. For instance, both the Original and DFKD demonstrate MIA$_{II}$ values of approximately 40\% or greater, which are markedly higher than those observed in the Retrain. In the comparison methods, BlockF's MIA$_{II}$ is typically higher than Retrain's and nearly equivalent to Original's, indicating that there is still a considerable amount of knowledge related to forgetting data in the unlearned model obtained by BlockF. GKT has the lowest MIA$_{II}$, which is because GKT is unable to effectively maintain the knowledge of retaining classes, resulting in the unlearned model's overall performance being low, and more akin to a model that has not been adequately trained. As previously stated, the signal generated by an inadequately trained model can also confuse the attacker, resulting in a low success rate of the attack. Therefore, the MIA$_{II}$ metric should be as close to Retrain as possible. The results demonstrate that our proposed method's MIA$_{II}$ is the closest to Retrain.

\textbf{3) AIN: }
Models with complete knowledge have AIN scores close to or even less than 0.1, as observed in those of DFKD and BlockF. This means that the models obtained by these two methods require only a few training steps to restore $A_f$ performance to a level comparable to that of the original model and also indicates that these models still contain forgetting related knowledge. When the AllCNN network is employed, our proposed methods yield AIN scores of approximately 1, indicating that the unlearned models obtained through our proposed methods contain an equivalent amount of forgetting-related knowledge as the retrained model, which is essentially negligible. While the AIN value of our proposed method is smaller when using the ResNet18, it remains considerably higher than that of DFKD and BlockF, still indicating that the unlearned models acquired through \methodname~incur a substantially higher cost to restore $A_f$ to a comparable level to original model.

\subsubsection{Efficiency}
\chenhao{To demonstrate the efficiency of \methodname, we report the time taken by each method to achieve a similar $A_r$ in the first column of Figure~\ref{fig:efficiency}. Since \postfilter~and \prefilter~process different numbers of samples during distillation, the time required for each training step varies. Therefore, we use wall time (actual elapsed time) for a more fair comparison of their efficiency. The results show that \methodname~requires significantly less training time than GKT to reach the same $A_r$, demonstrating substantial efficiency gains for both \postfilter~and \prefilter~techniques within \methodname.}

\subsection{Ablation}

In this subsection, we conduct ablation experiments on two key techniques in the \methodname~to elucidate their contributions respectively. Specifically, when examining the \inhibsynShort, we utilize the \prefilter~to filter forgetting-related knowledge; when examining the \postfilterShort, we just exclude the \inhibsynShort~from \methodname.

\subsubsection{How does the \inhibsyn~(\inhibsynShort) work?}
As previously discussed in the Section~\ref{sec:anayzeAdv}, in the unlearning scenario, in the absence of constraints during the generator's training process, the generator will progressively synthesize more samples that contain information about forgetting classes. As illustrated in the second column in Figure~\ref{fig:efficiency}, in methods that do not incorporate the \inhibsynShort, such as GKT and \postfilterShort, an increasing number of samples comprising information about forgetting classes are synthesized as training progresses, whereas a decreasing number of samples contain information about retaining classes. Furthermore, the third column of Figure~\ref{fig:efficiency} demonstrates that the number of samples filtered out by the \prefilter~is considerably larger than the number of samples belonging to the forgetting classes. This suggests that a significant proportion of samples from the retaining classes are also filtered out due to their resemblance to the forgetting classes, which further reduces the number of samples involved in distillation and reduces the student's learning efficiency. By comparing the results of GKT and \inhibsynShort, \inhibsynShort~not only markedly reduces the synthesis of samples for forgetting classes, but also significantly reduces the number of samples that are filtered out by the \prefilter. This can also be seen from the representation visualization in Figure~\ref{fig:testsetVSgenset}. This suggests that \inhibsynShort~is an effective method of suppressing the synthesis of samples that contain forgetting class information, thereby markedly enhancing student's learning efficiency in retaining knowledge.

\subsubsection{How does the \postfilter~(\postfilterShort) work?}
A comparison of the results of GKT and \postfilterShort~reveals that, as neither employs \inhibsynShort~to restrict the learning of the generator, both appear to synthesize an increasing number of forgetting class samples as the training progresses. However, in contrast to GKT, the \postfilterShort~does not remove any sample before distillation. Instead, it utilizes as much information as possible about the retaining classes in all samples. To this end, we quantify the information entropy of the retaining classes in the output of the teacher, in each training step. Specifically, we calculated the information entropy in each training step as
\begin{equation}
    H_B=-\sum_i^N\sum_{k\notin\mathcal{Y}_f}T_k(\widetilde{x}_i)\cdot \log(T_k(\widetilde{x}_i)),
\end{equation}
where the $N$ is the number of samples that are involved in distillation. We report the average $H_B$ across all training steps in the fourth column in Figure~\ref{fig:efficiency}. As shown in the figure, \postfilterShort~can provide a considerably greater quantity of information to the student in each training step when compared to the GKT. This suggests that, although the \postfilterShort~also encounters the challenge of an increasing number of synthetic forgetting class samples, its utilization of the retaining information within the samples to its fullest potential also markedly enhances the learning efficiency of the student.

\subsection{Additional analysis}

\subsubsection{Visualization}
\begin{figure}[t]
\centering
    \subfigure[GKT]{
        \begin{minipage}[t]{0.45\linewidth}
            \centering
            \includegraphics[width=\linewidth]{imgs/repre_vis_orig_test_vs_gen_svhn_allcnn_dfq_GKT_7.pdf}
        \end{minipage}
        \label{fig:testsetVSgenset_GKT}
    }
    \hfill
    \subfigure[\inhibsynShort]{
        \begin{minipage}[t]{0.45\linewidth}
            \centering
            \includegraphics[width=\linewidth]{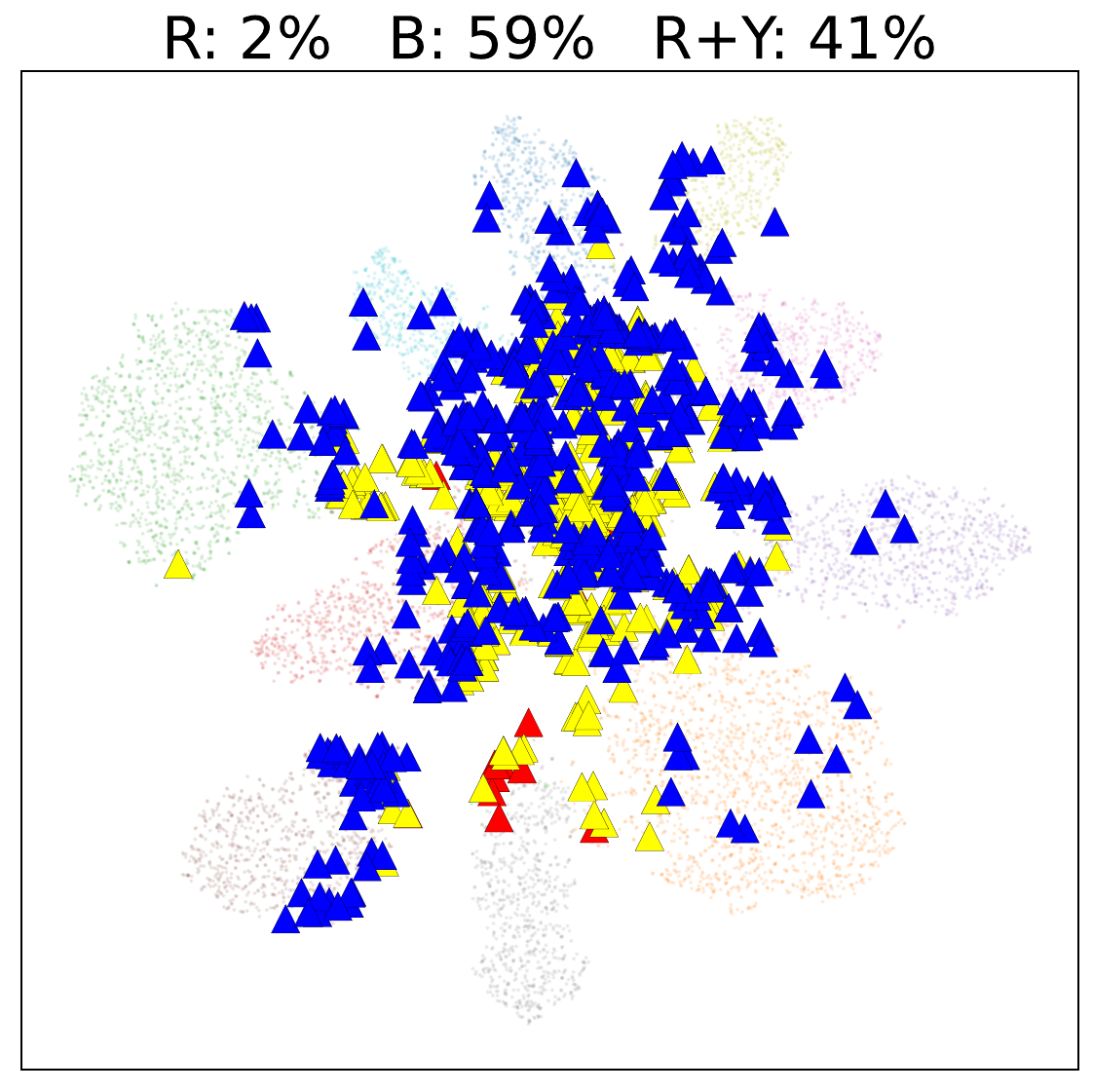}
        \end{minipage}
        \label{fig:testsetVSgenset_GKTFA}
    }

    \subfigure[\postfilterShort]{
        \begin{minipage}[t]{0.45\linewidth}
            \centering
            \includegraphics[width=\linewidth]{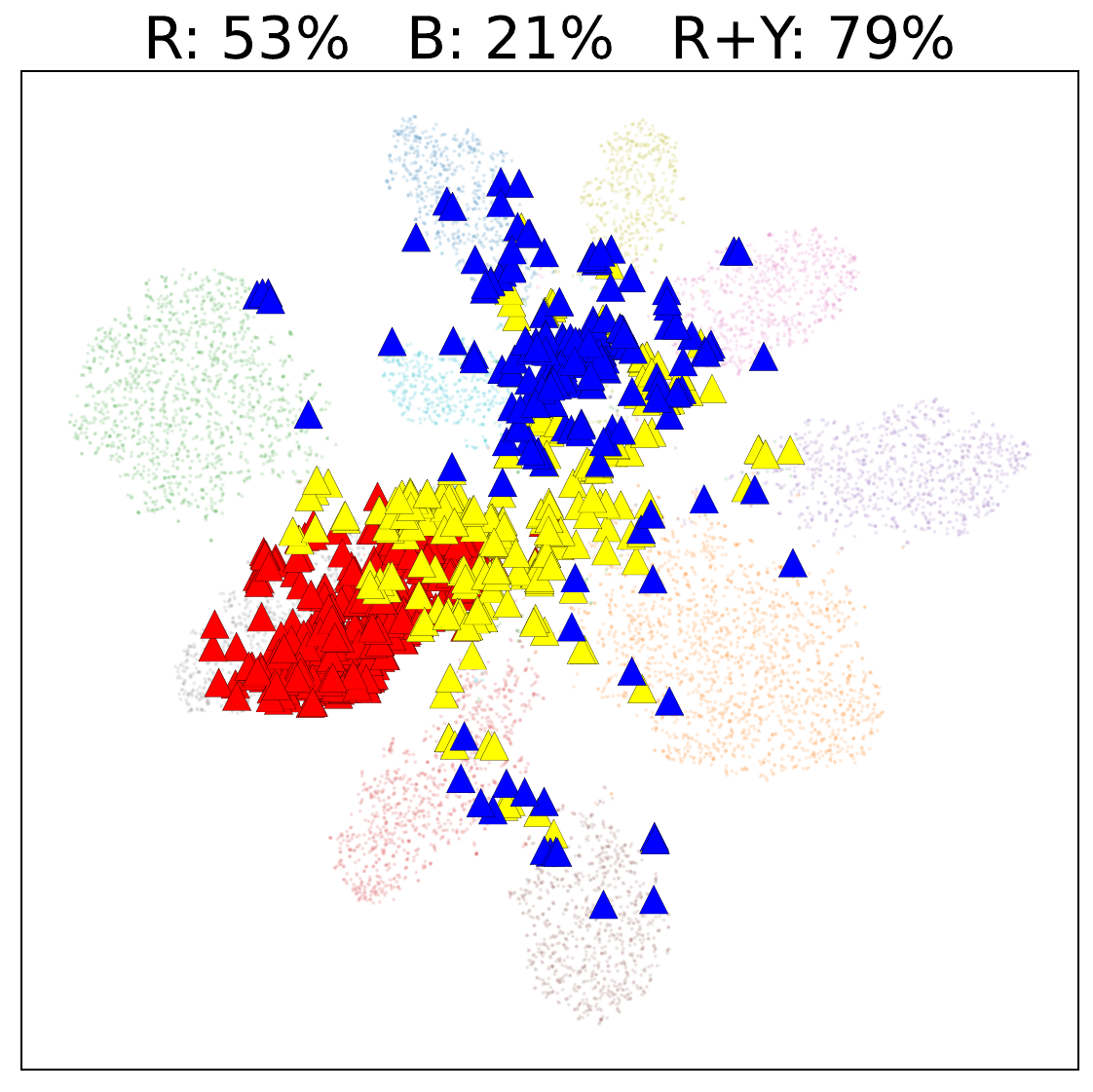}
        \end{minipage}
        \label{fig:testsetVSgenset_PostFilter}
    }\hfill
    \subfigure[\methodname]{
        \begin{minipage}[t]{0.45\linewidth}
            \centering
            \includegraphics[width=\linewidth]{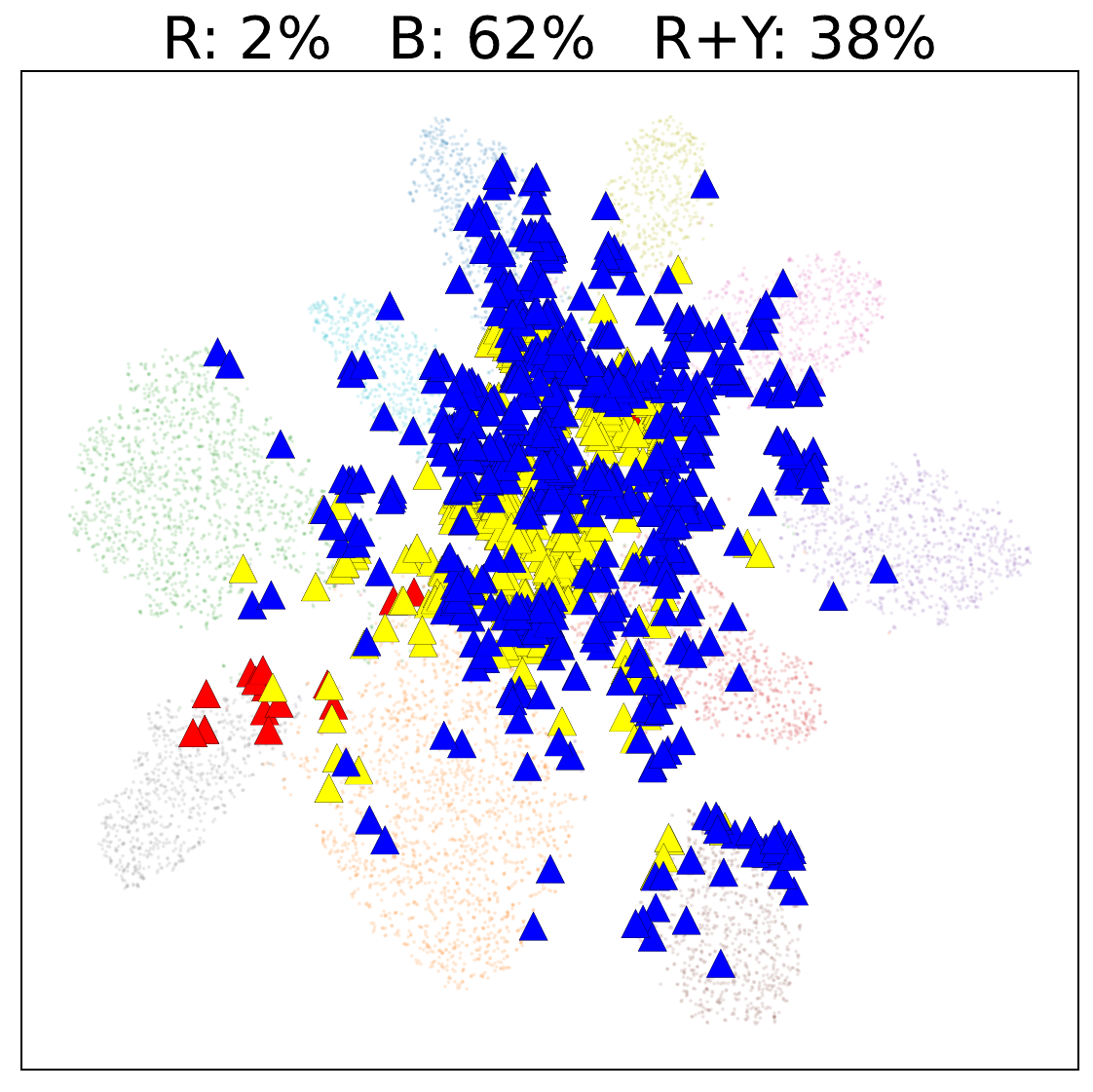}
        \end{minipage}
        \label{fig:testsetVSgenset_PostFilterFA}
    }

\caption{Visualization on SVHN.}
\label{fig:testsetVSgenset}
\end{figure}

We visualize the representation of synthetic samples during the experiments on SVHN, using the unlearning of ``7" as an example, and the results are plotted in Figure~\ref{fig:testsetVSgenset}. The symbols in Figure~\ref{fig:testsetVSgenset} are consistent with those in Figure~\ref{fig:problem}. Compared to GKT, our \inhibsynShort, while still using \prefilter, significantly suppresses the synthesis of the forgetting class samples and increases the number of samples involved in the distillation. While the number of synthesized forgetting class samples is not reduced in \postfilterShort, it makes full use of retaining-related information from all synthetic samples. When \postfilterShort~is combined with \inhibsynShort, which is the \methodname, the number of synthesized samples containing forgetting information is further reduced and all samples are fully utilized. We also visualize the synthetic images in Appendix~\ref{sec:synSample}.

\subsubsection{CIFAR-100 Results}
\begin{table}[t]
\centering
\setlength\tabcolsep{1pt}
\begin{tabular}{c|cc|cc}
\toprule
\multirow{2}{*}{Method} & \multicolumn{2}{c|}{Single-Class} & \multicolumn{2}{c}{Multi-Classes} \\
                        & $A_f$           & $A_r$           & $A_f$          & $A_r$          \\
\midrule
Original                & 70.3$\pm$0.3    & 70.31$\pm$0.33              & 70.3$\pm$0.3   & 70.31$\pm$0.33              \\
DFKD                    & 61.8$\pm$0.1    & 61.81$\pm$0.16              & 61.8$\pm$0.1   & 61.8$\pm$0.16              \\ \cline{1-5}
Retrain                 & 0.0$\pm$0.0     & 70.36$\pm$0.11              & 0.0$\pm$0.0    & 70.13$\pm$0.04              \\ 
GKT                     & 0.0$\pm$0.0     & 59.65$\pm$0.37              & 0.0$\pm$0.0    & 49.86$\pm$3.01              \\
\inhibsynShort~(ours)                      & 0.0$\pm$0.0     & 60.72$\pm$0.36              & 0.0$\pm$0.0    & \underline{58.32$\pm$0.51}              \\
\postfilterShort~(ours)              & 0.0$\pm$0.0     & \underline{61.67$\pm$0.36}  & 0.0$\pm$0.0    & 57.07$\pm$2.19              \\
\methodname~(ours)           & 0.0$\pm$0.0     & \textbf{62.14$\pm$0.25}     & 0.0$\pm$0.0    & \textbf{62.58$\pm$0.62}              \\
\bottomrule
\end{tabular}
\caption{CIFAR-100 results.}
\label{tab:cifar100Res}
\end{table}
As shown in Table~\ref{tab:cifar100Res}, \methodname~still outperforms the existing method, particularly in the setting of multi-classes unlearning.
For multi-classes unlearning, we select class labels $\mathcal{Y}_f=\{0,1,2,3,4,5,6,8,9,12\}$. Note that this is a more difficult setup. If the chosen forgetting classes are sharing one superclass, the task can be reduced to forgetting one superclass and fewer retaining classes will be filtered out by the \prefilter.
However, if the selected forgetting classes are spread across superclasses, samples from other retaining classes in each superclass will be filtered out due to their similarity to the forgetting classes, resulting in more samples being filtered out in a batch. We also chose 10 classes from two superclasses and the results (Appendix~\ref{sec:concenMultiClasses}) show that the number of samples involved in the distillation in GKT is significantly more than that when 10 classes are distributed over 10 superclasses, and the final result of $A_r=56.04$ (vs. $A_r=49.86$ in Table~\ref{tab:cifar100Res}) is also better. This result demonstrates that the \methodname~exhibits superior performance in a more challenging setting, even when compared to the existing method that operates under an easier setting.

\section{Conclusion}
We theoretically analyzed and experimentally demonstrated the inefficiency in retaining knowledge during data-free unlearning when using Data-free Knowledge Distillation (DFKD). Our findings show that enriching the information related to retaining classes during distillation significantly enhances the student model's learning of retaining-related knowledge. We propose the Inhibited Synthetic PostFilter (ISPF) to achieve this from two perspectives: 1) reducing the synthesis of forgetting class information and 2) fully leveraging the retaining-related information in the synthesized samples. Experimental results confirm that ISPF effectively overcomes this challenge and outperforms existing methods.

\section*{Acknowledgements} The work is supported by the Australian Research Council Discovery Projects DE230101116 and DP240103070. 

\bibliography{aaai25}

\appendix

\section{\chenhao{Related Works}}
\label{sec:extendRelatedWorks}

\subsection{Machine Unlearning}
\chenhao{There are two principal categories of unlearning methodologies~\cite{unlearningSurvey1,unlearningSurvey2}: exact unlearning and approximate unlearning. The former~\cite{SISA,ARCANE,TwostageModelRetraining} usually entails excluding the data to be forgotten (forgetting data) from the training data set, followed by the retraining of a model on the retained data. These methods inevitably necessitate access to the original training data set. The latter is achieved by tuning the parameters of the trained model so that the performance of the tuned model approximates that of the retrained model. These approaches primarily use forgetting data for gradient ascent~\cite{unrolling} or modifying the most relevant network parameters~\cite{SalUN,SSD}, and retaining data to repair the performance degradation caused by modifying the parameters. All of these methods necessitate access to real data to identify learning objectives and achieve unlearning.}

\chenhao{Due to storage expenses and privacy concerns, real training data are often deleted or archived post-training, leading to scenarios where unlearning methods must operate without full access to the real dataset. UNSIR~\cite{UNSIR} eliminates access to forgetting data by learning proxy noise samples. Boundary Unlearning~\cite{boundaryUnlearning} requires only forgetting data and tuning the model with relabeled forgetting data. In contrast to these methods, our method does not require access to real data, neither forgetting nor retaining data.}

\subsection{Data-free Knowledge Distillation \& Data-Free Unlearning}
\chenhao{To achieve distillation without real training data, which is a more common scenario in practice, a branch of research called Data-free Knowledge Distillation (DFKD)~\cite{DFAD,FastDFKD,ZSKD,DFKDDream,DFKD_DI} emerged. Among existing top-performance DFKD methods~\cite{DAFL,DFQ,ZSKD,adv-DFKD,SSDFKD,FastDFKD,CMI}, the adversarial inversion-and-distillation paradigm has been widely adopted, where a generator is trained to produce pseudo-samples opposing the student's learning process. Specifically, the generator aims to create data that the student model struggles to correctly learn or classify, effectively working against the student's learning objective }

\chenhao{Chundawat et al.~\cite{GKT} first raised the Zero-shot Unlearning problem (referred to as Data-free Unlearning in this work), in which the unlearning method can only access the trained original model and the forgetting class label. They proposed Gated Knowledge Transfer (GKT) that applied one of the DFKD methods under adversarial inversion-and-distillation paradigm ~\cite{ZSKD} for generating samples and designed a filter to select generated samples whose likelihood of belonging to the forgetting class is below a specified threshold. Although it is a pioneering work in exploring Data-free Unlearning, GKT displays less efficiency in maintaining retained knowledge. In this work, we are inspired by the DFKD to obtain unlearned models by transferring only the retained knowledge from the original model to a new network. We revisit the challenges that DFKD encounters when applied to the unlearning problem, analyze the underlying reason for GKT's inefficiency, and propose a more efficient method for Data-free Unlearning.}

\section{\methodname~Algorithm}
\label{sec:algo}
The overall algorithm of \methodname~is illustrated in the Algorithm~\ref{alg:ISPF}. The variables $n_g$ and $n_s$ represent the number of training steps for the generator and student model, respectively. Similarly, $\eta_g$ and $\eta_s$ denote the learning rates for the generator and student models, respectively. To represent the update of parameters, we extend the notations of the generator and student models to $G(\phi,\cdot)$ and $S(\theta,\cdot)$, where the $\phi$ and the $\theta$ are their parameters, respectively. In the context of unlearning, the trained student $S(\theta,\cdot)$ is the unlearned model.

\begin{algorithm}[hbt!]
\caption{\inhibsyn~\postfilter~(\methodname)}\label{alg:ISPF}
\begin{algorithmic}
\REQUIRE{~\\
Pretrained original model $T(\cdot)$\\
Init generator $G(\phi,\cdot)$\\
Init student $S(\theta,\cdot)$}\\
Training parameters $n_g$, $n_s$, $\eta_g$ and $\eta_s$
\ENSURE{the trained $S(\theta,\cdot)$ as the unlearned model.}
\FOR{in $range(epochs)$}

 \FOR{in $range(loops)$}
    \STATE \# generator steps
    \FOR{1,2,...,$n_g$}
        \STATE $z\sim\mathcal{N}(\boldsymbol{0},\boldsymbol{I})$
        \STATE $\widetilde{x}\leftarrow G(\phi,z)$
        \STATE $L_g\leftarrow \mathcal{L}_{IS}(\widetilde{x})$\; (Eq.~\ref{eq:IA})
        \STATE $\phi\leftarrow \phi-\eta_g\frac{\partial L_g}{\partial\phi}$
     \ENDFOR
     \STATE \# student steps
     \FOR{1,2,...,$n_s$}
        \STATE $z\sim\mathcal{N}(\boldsymbol{0},\boldsymbol{I})$
        \STATE $\widetilde{x}\leftarrow G(\phi,z)$
        \STATE $L_s\leftarrow \mathcal{L}_{S}(\widetilde{x})$\; (Eq.~\ref{eq:KD_ReD})
        \STATE $\theta\leftarrow \theta-\eta_s\frac{\partial L_s}{\partial\theta}$
     \ENDFOR
 \ENDFOR
 \STATE decay $\eta_s$
\ENDFOR
\end{algorithmic}
\end{algorithm}

\section{Implementation Details}
\label{sec:hyperparam}
We train the original model with the following hyperparameters and apply a MultiStepLR scheduler with a gamma of 0.1. The \textit{Retrain} method shares the same setting with the training of the original model.
\begin{table}[ht]
\centering
\begin{tabular}{ccccc}
\toprule
         & lr   & epoch & batch-size & lr decay at \\
\midrule
AllCNN   & 0.01 & 10    & 256        & -           \\
ResNet18 & 0.01 & 10    & 256        & -           \\
\bottomrule
\end{tabular}
\caption{Original model training setting for SVHN.}
\label{tab:orig_svhn}
\end{table}

\begin{table}[ht]
\centering
\begin{tabular}{ccccc}
\toprule
         & lr   & epoch & batch-size & lr decay at \\
\midrule
AllCNN   & 0.1 & 50     & 256        & 30           \\
ResNet18 & 0.1 & 100    & 256        & 60,80           \\
\bottomrule
\end{tabular}
\caption{Original model training setting for CIFAR-10.}
\label{tab:orig_cifar10}
\end{table}

\begin{table}[ht]
\centering
\begin{tabular}{ccccc}
\toprule
         & lr   & epoch & batch-size & lr decay at \\
\midrule
ResNet18 & 0.1 & 100    & 256        & 60,80           \\
\bottomrule
\end{tabular}
\caption{Original model training setting for CIFAR-100.}
\label{tab:orig_cifar100}
\end{table}

For methods that use the \prefilter, the confidence threshold $\delta$ is a hyperparameter. According to the GKT~\cite{GKT}, the threshold should be less than or equal to $1/K$ where $K$ is the number of classes. The exact threshold for ten-class datasets in the GKT is 0.01, we therefore set the $\delta=0.01$ for SVHN and CIFAR-10 and $\delta=0.001$ for CIFAR-100.

For the setting of DFKD, we refer to the original setting in the DFQ~\cite{DFQ} as well as one of its reproduction in~\cite{FastDFKD} and set a learning rate of $1e^{-3}$ and a synthesis batch size of 256 for all experiments. For the SVHN, the distillation epoch number is 100. During each epoch, a single loop of student and generator training is conducted. Within this loop, the student is trained for 10 steps, while the generator is trained for 1 step. The student's learning rate is 0.1 when using AllCNN and 0.05 when using ResNet18. For the CIFAR-10, the distillation epoch number is 200. During each epoch, 5 loops of student and generator training are conducted. In each loop, the student is trained for 10 steps, while the generator is trained for 1 step. The student's learning rate is 0.05 for both AllCNN and ResNet18. CIFAR-100 employs the same configuration as that utilized in CIFAR-10, except for the loop number, which is 10. For SVHN and CIFAR-10 datasets, experiments were conducted with each class as a forgetting class separately. All experiments are repeated for three trials with different random seeds.

All the experiments are conducted on one server with NVIDIA RTX A5000 GPUs (24GB GDDR6 Memory) and 12th Gen Intel Core i7-12700K CPUs (12 cores and 128GB Memory). The code was implemented in Python 3.9.12 and Cuda 12.0. The main Python packages' versions are the following: Numpy 1.22.4; Pandas 1.4.3; Pytorch 1.12.1; Torchvision 0.13.1. 

\section{Results when Unlearning Each Class}
\label{sec:detailRes}

\begin{figure}[th]
\centering
    \subfigure[SVHN-AllCNN]{
        \begin{minipage}[t]{0.45\linewidth}
            \centering
            \includegraphics[width=\linewidth]{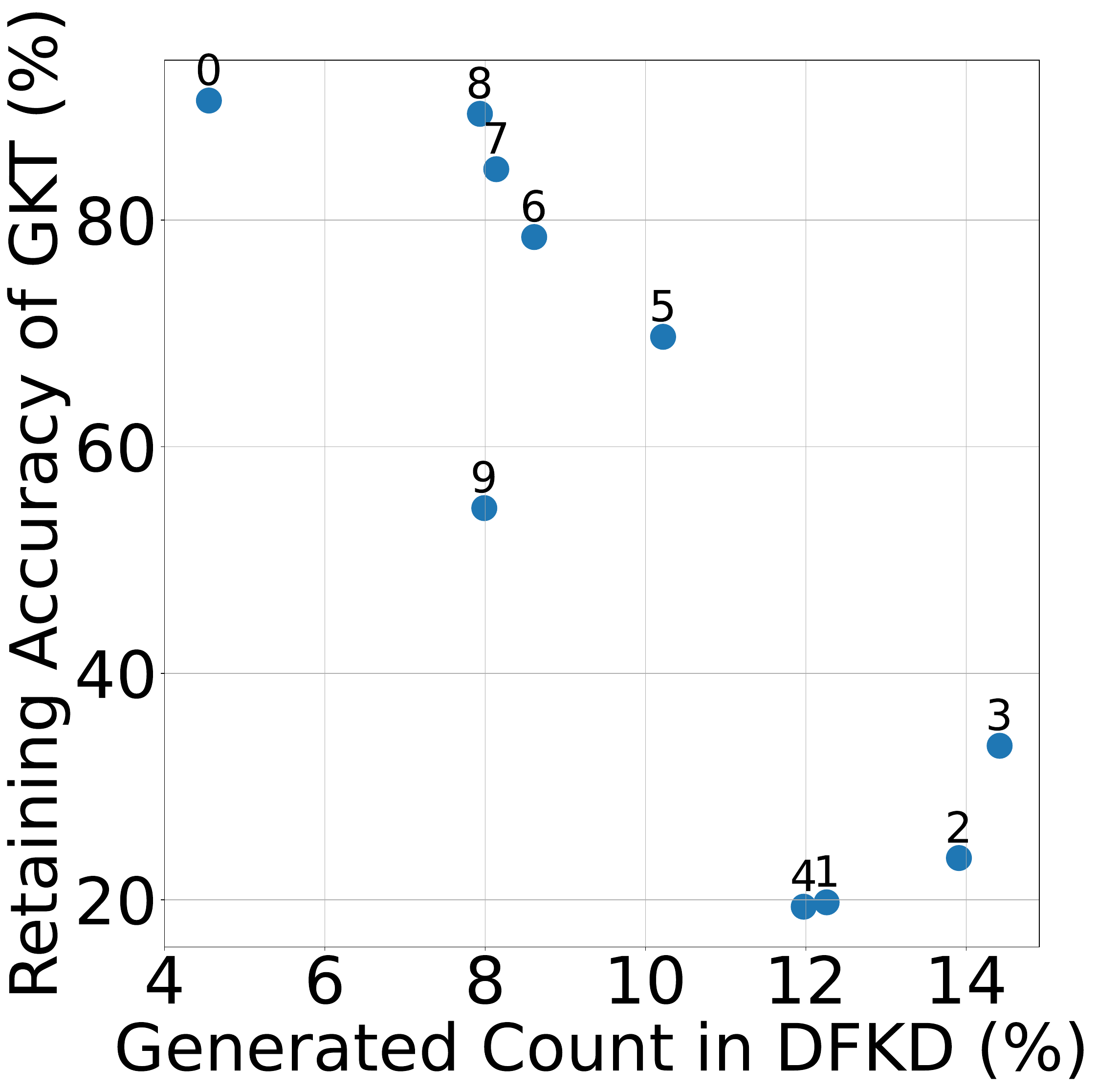}
        \end{minipage}
        \label{fig:DFKDgen_GKTacc_svhn_allcnn}
    }
    \hfill
    \subfigure[SVHN-ResNet18]{
        \begin{minipage}[t]{0.45\linewidth}
            \centering
            \includegraphics[width=\linewidth]{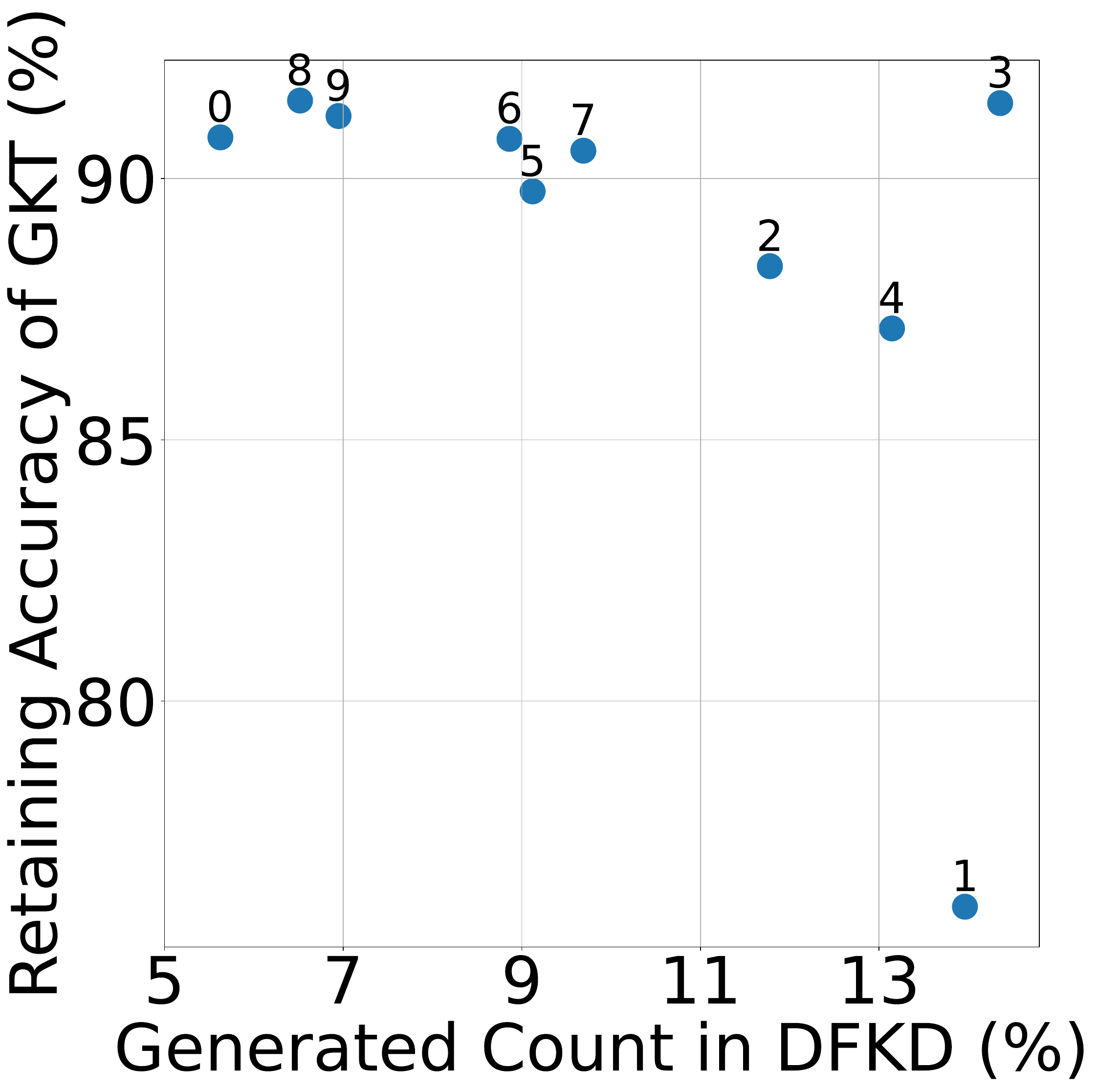}
        \end{minipage}
        \label{fig:DFKDgen_GKTacc_svhn_resnet18}
    }

    \subfigure[CIFAR10-AllCNN]{
        \begin{minipage}[t]{0.45\linewidth}
            \centering
            \includegraphics[width=\linewidth]{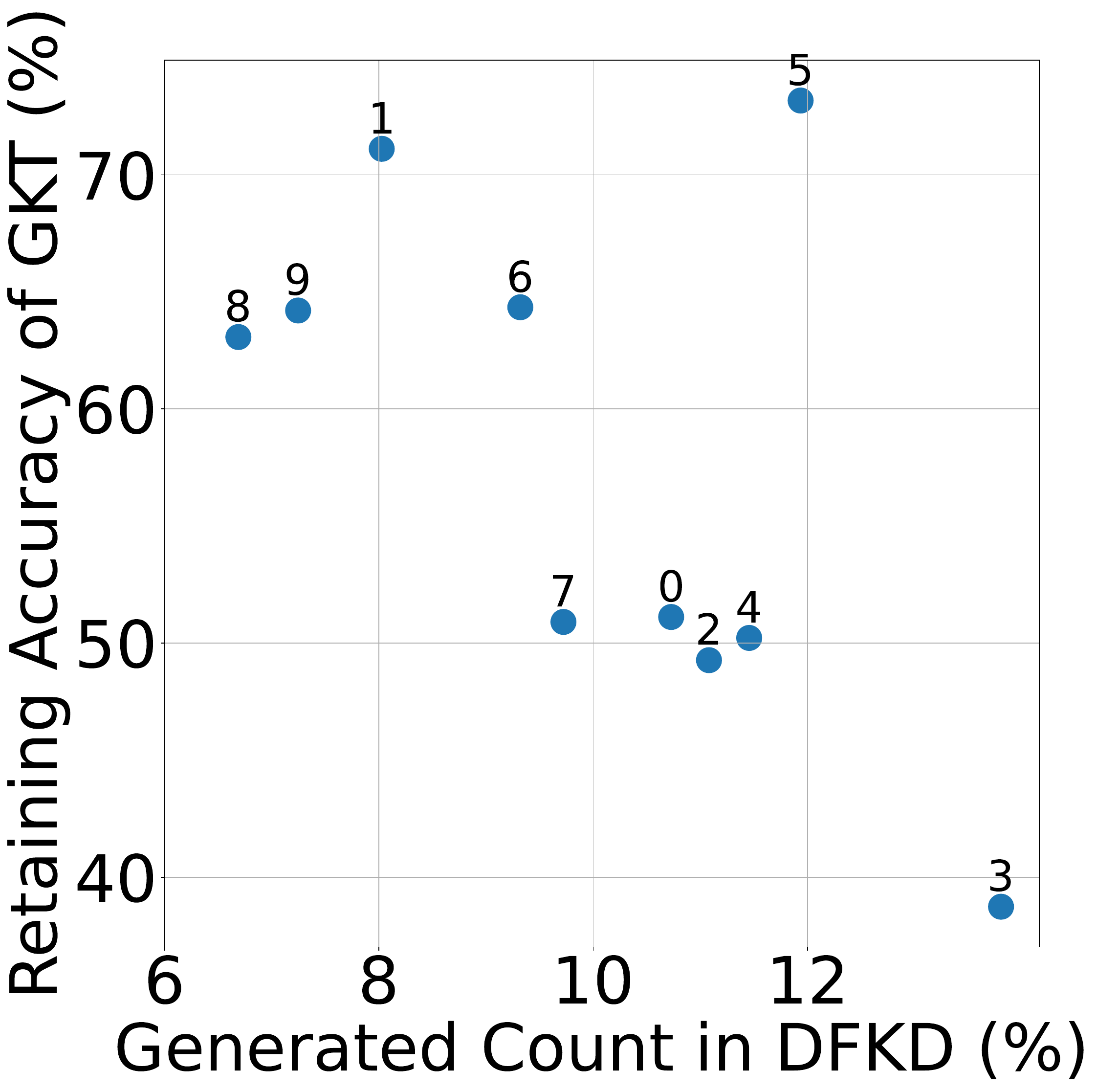}
        \end{minipage}
        \label{fig:DFKDgen_GKTacc_cifar10_allcnn}
    }
    \hfill
    \subfigure[CIFAR10-ResNet18]{
        \begin{minipage}[t]{0.45\linewidth}
            \centering
            \includegraphics[width=\linewidth]{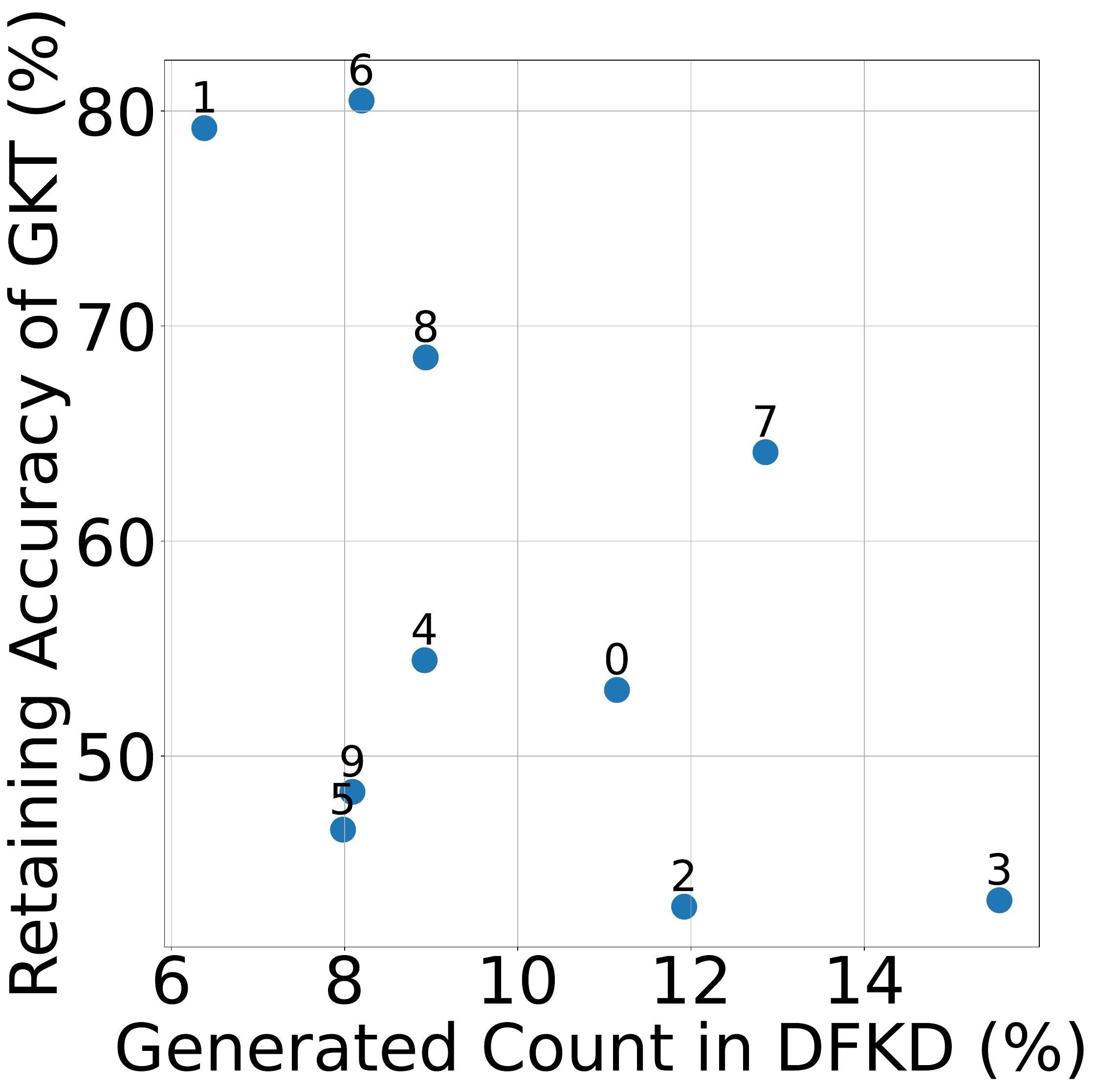}
        \end{minipage}
        \label{fig:DFKDgen_GKTacc_cifar10_resnet18}
    }

\caption{Difficulty of synthesizing samples for each class vs. the $A_r$ of the GKT. The x-axis represents the proportion of samples synthesized by the pure DFKD generator for each class. This value is used to reflect the difficulty of synthesizing samples for a given class, i.e., a higher proportion corresponds to a lower difficulty. The number on the top of each point is the class index.}
\label{fig:DFKDgen_vs_GKTacc}
\end{figure}

In Table~\ref{tab:detailRes}, we show in detail the performance results of the unlearned model obtained by each method when each individual class is used as the forgetting class. We can see that GKT performs differently when forgetting different classes. For example, GKT performs well when unlearning ``0" under the SVHN-AllCNN setting, but when unlearning ``1", the model's performance on retaining classes is severely damaged. We believe that this is because the generator has a different difficulty in exploring the distribution of different classes which phenomenon has also been discussed in \cite{SSDFKD}. To demonstrate the influence of the generator's varying capacity to synthesize samples for classes on the performance of the existing data-free unlearning method, we treat the proportion of samples synthesized for each class in the pure DFKD as a measure of synthesis difficulty, i.e., the more samples synthesized for a class, the easier the class is to be explored by the generator. We observe the synthesis difficulty of each class versus the performance of GKT, and present the results in Figure~\ref{fig:DFKDgen_vs_GKTacc}. The results demonstrate that classes with a higher proportion of synthesized samples in the pure DFKD tend to exhibit lower $A_r$ in the unlearning results of the GKT. This is attributed to the fact that these classes are more readily for the generator to explore and synthesize, and are also more likely to be significantly filtered out when they are unlearned by the GKT, thus reducing the subsequent distillation efficiency and $A_r$. In contrast, the performances of our proposed method are consistent across classes. Meanwhile, our proposed methods outperform the baselines, especially the best performance achieved when \postfilter~and \inhibsyn~are combined.

\begin{figure*}[t]
    \subfigure[SVHN-ResNet18]{
        \begin{minipage}[t]{\linewidth}
            \centering
            \includegraphics[width=\linewidth]{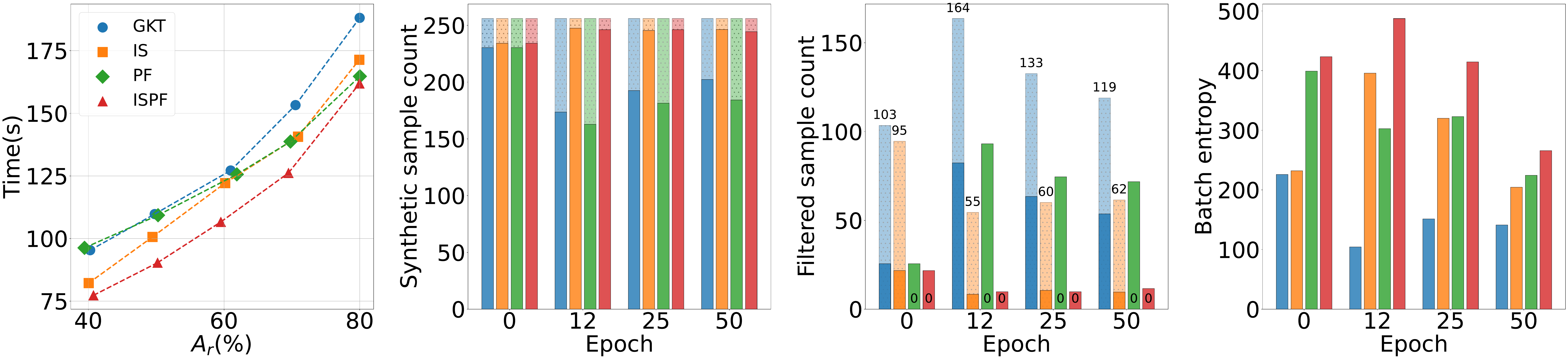}
        \end{minipage}
        \label{fig:svhnResnet18}
    }
    \subfigure[CIFAR10-ResNet18]{
        \begin{minipage}[t]{\linewidth}
            \centering
            \includegraphics[width=\linewidth]{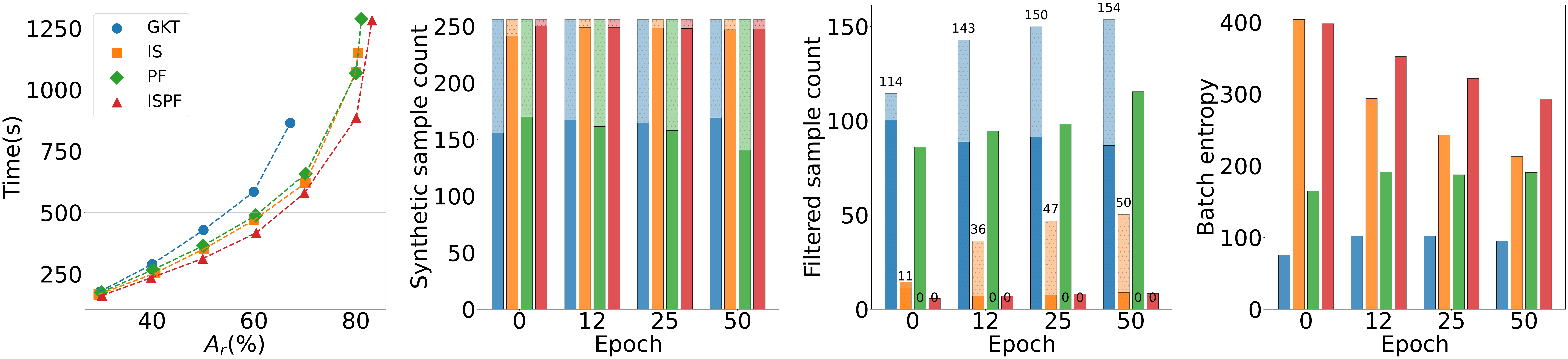}
        \end{minipage}
        \label{fig:cifar10Resnet18}
    }

\caption{The colors in all figures are used to distinguish the different methods, and each method is shown with its corresponding color in the legend of the first column. The top row shows the results under the SVHN-ResNet18 setting, and the bottom row shows the results under CIFAR10-ResNet18. The \textbf{first column} shows the results for $A_r$ vs. wall time. In the \textbf{second column}, the light bar filled with dots shows the number of synthetic samples classified as forgetting classes by the original model, and the dark bar shows the number of retaining class samples. In the \textbf{third column}, the light bar filled with dots shows the number of samples filtered out before distillation and the digits indicate the exact number of filtered-out samples, and the darker bar shows the number of synthetic forgetting class samples, which is the same as the lighter bar in the second column.}
\label{fig:addEfficiency}
\end{figure*}

\section{Detailed Results on ResNet18}
\label{sec:resnet18Res}
In the experimental analysis subsection of the main text, we only presented results for the AllCNN network structure. In this appendix, we report the corresponding results for ResNet18. The results are shown in Figure~\ref{fig:addEfficiency}, where all notations are consistent with Figure~\ref{fig:efficiency} in the main text. We can still get consistent conclusions from the results as stated in the main text.

\section{Shared Multi-classes in CIFAR-100}
\label{sec:concenMultiClasses}

To demonstrate the difference between forgetting classes that are similar to each other and those that are not similar in multi-class unlearning experiments, we selected 10 classes ($\mathcal{Y}_f=\{4,30,55,72,95,1,32,67,73,91\}$) from two superclasses (``aquatic mammals'' and ``fish'') in CIFAR-100 for analysis. For the sake of clarity, we abuse the term ``\textit{Shared}'' to refer to the setting of $\mathcal{Y}_f=\{4,30,55,72,95,1,32,67,73,91\}$ and use the term ``\textit{Distributed}'' to refer to the setting of $\mathcal{Y}_f=\{0,1,2,3,4,5,6,8,9,12\}$.

As evidenced by Table~\ref{tab:concentratedMultiClasses}, each method in the \textit{Shared} setting demonstrates superior performance compared to the \textit{Distributed} setting. The methods employing \prefilter~have fewer synthetic samples that are filtered out under the \textit{Shared} setting, thereby enhancing the quantity of information utilized for distillation. Furthermore, our methods exhibit better performance under the \textit{Distributed} setting compared to the existing method under the easier \textit{Shared} setting.

\begin{table}[ht]
\centering
\setlength\tabcolsep{1.5pt}
\begin{tabular}{c|cc|cc}
\toprule
\multirow{2}{*}{Method} & \multicolumn{2}{c|}{Shared} & \multicolumn{2}{c}{Distributed} \\
                        & \#Sample           & $A_r$           & \#Sample          & $A_r$          \\
\midrule
GKT                     & 101     & 56.04$\pm$1.02              & 68     & 49.86$\pm$3.01              \\
\inhibsynShort                      & 166     & \underline{60.44$\pm$0.59}  & 151    & \underline{58.32$\pm$0.51}              \\
\postfilterShort              & 256     & 58.91$\pm$1.44              & 256    & 57.07$\pm$2.19              \\
\methodname           & 256     & \textbf{63.74$\pm$0.58}     & 256    & \textbf{62.58$\pm$0.62}              \\
\bottomrule
\end{tabular}
\caption{Shared multi-classes vs. distributed multi-classes.}
\label{tab:concentratedMultiClasses}
\end{table}

\begin{figure*}[t]
    \subfigure[Unlearned model's predictive label distribution on $\mathcal{D}_f$ vs. that of the \textit{Retrain} model.]{
        \begin{minipage}[t]{\linewidth}
            \centering
            \includegraphics[width=\linewidth]{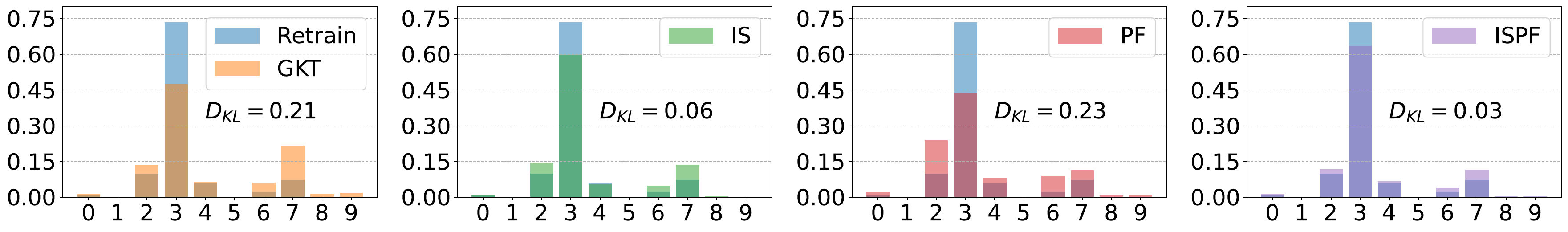}
        \end{minipage}
        \label{fig:predsDistCifar10_uncls_5}
    }
    \subfigure[GKT]{
        \begin{minipage}[t]{0.23\linewidth}
            \centering
            \includegraphics[width=\linewidth]{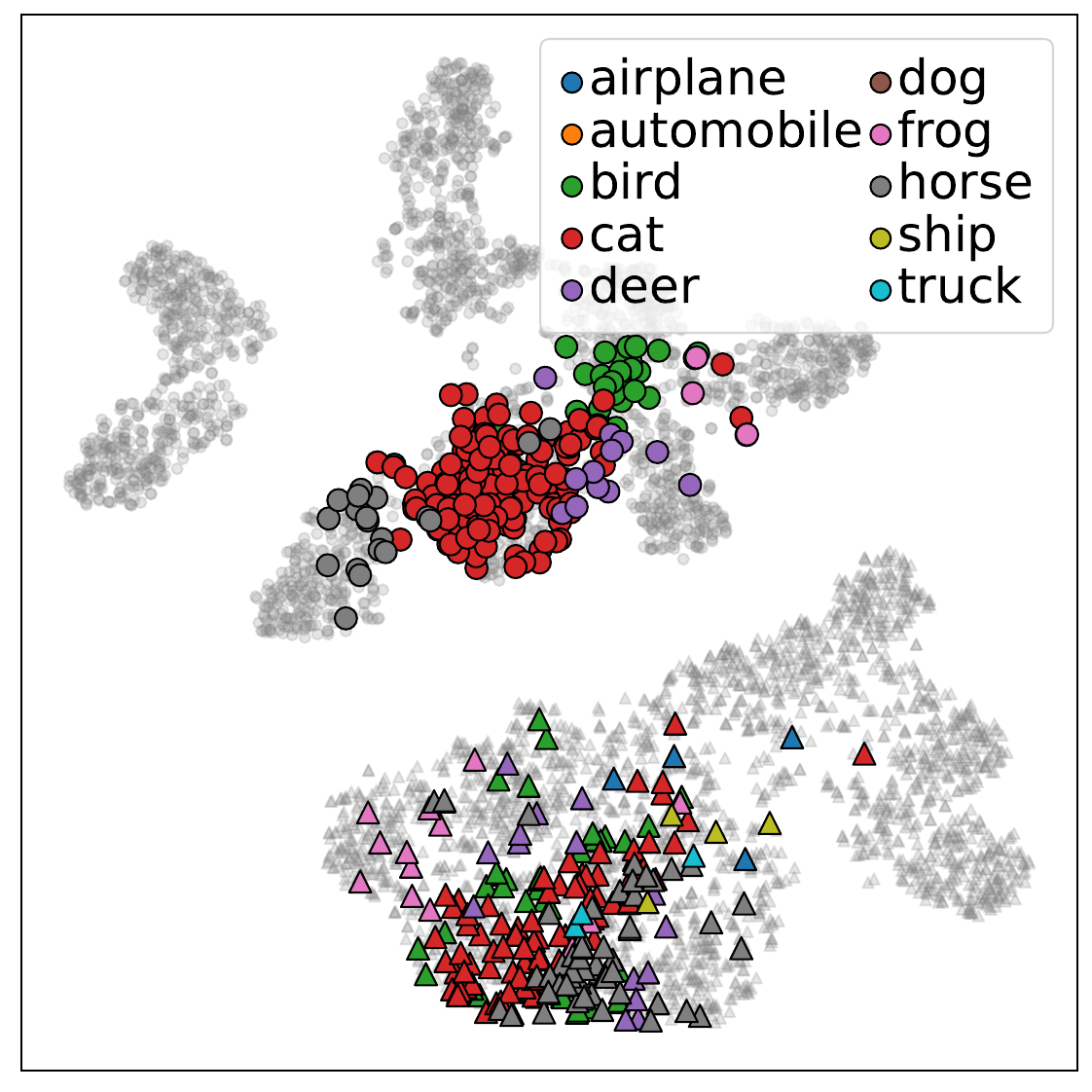}
        \end{minipage}
        \label{fig:predVis_GKT}
    }\hfill
    \subfigure[\inhibsynShort]{
        \begin{minipage}[t]{0.23\linewidth}
            \centering
            \includegraphics[width=\linewidth]{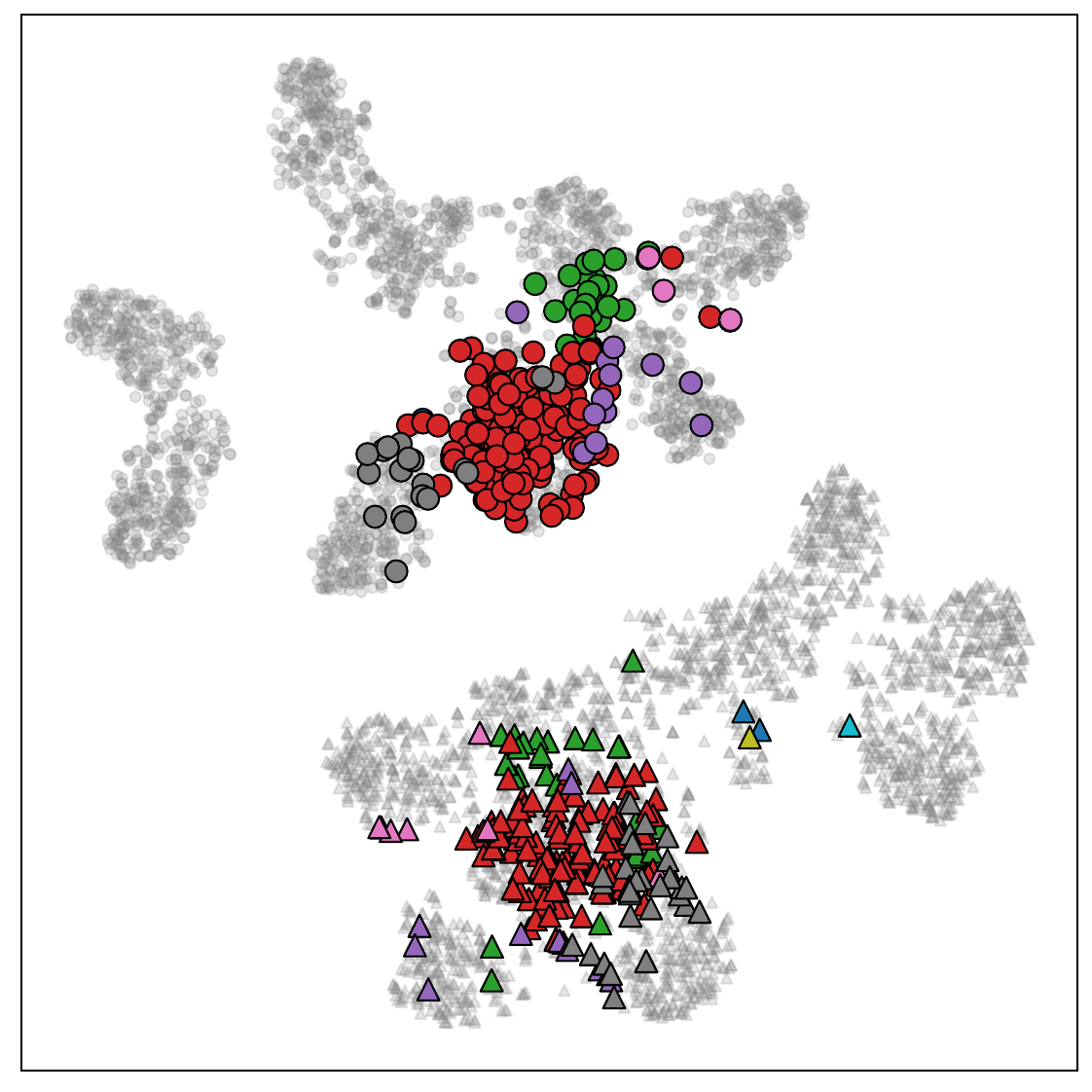}
        \end{minipage}
        \label{fig:predVis_IS}
    }
    \subfigure[\postfilterShort]{
        \begin{minipage}[t]{0.23\linewidth}
            \centering
            \includegraphics[width=\linewidth]{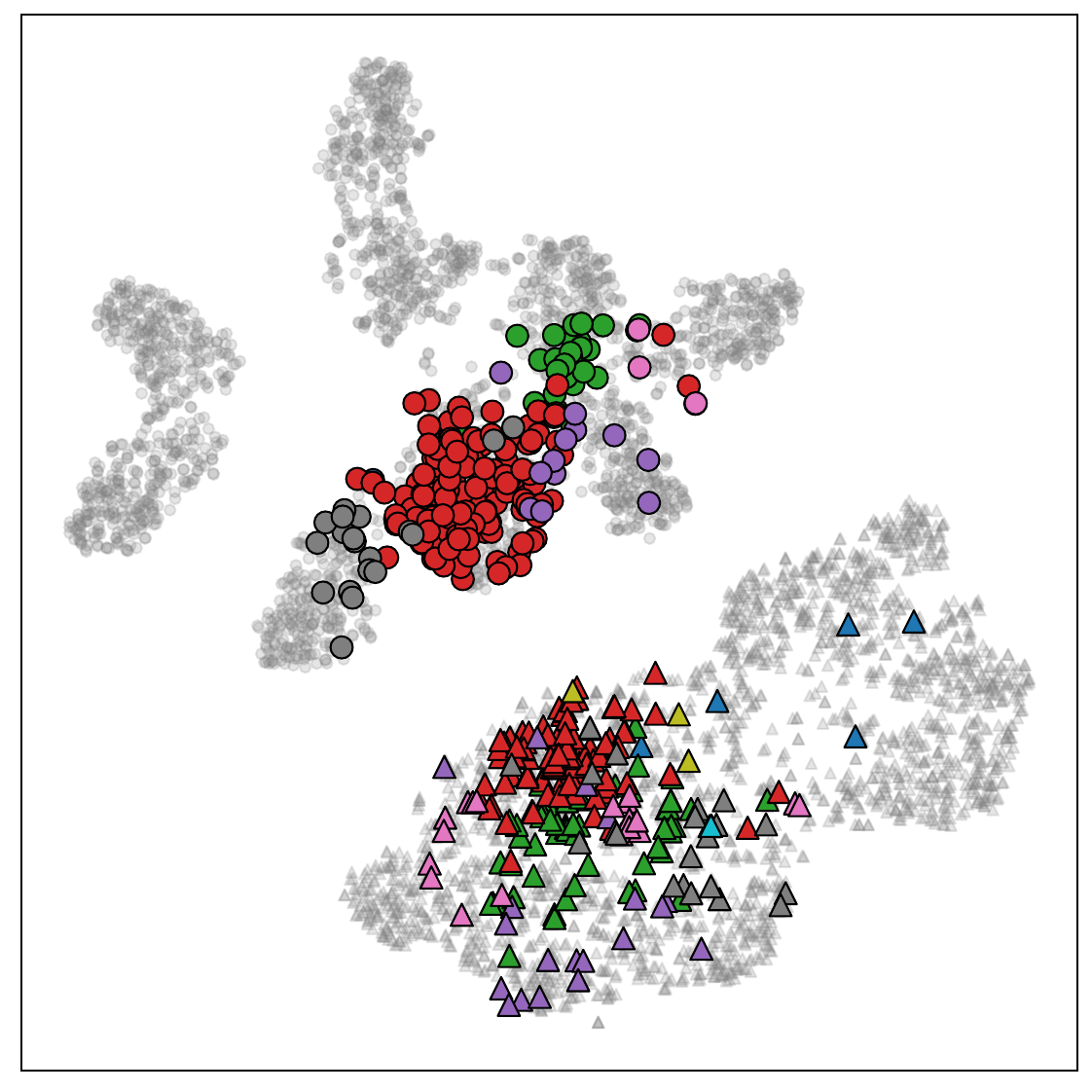}
        \end{minipage}
        \label{fig:predVis_PostF}
    }
    \subfigure[\methodname]{
        \begin{minipage}[t]{0.23\linewidth}
            \centering
            \includegraphics[width=\linewidth]{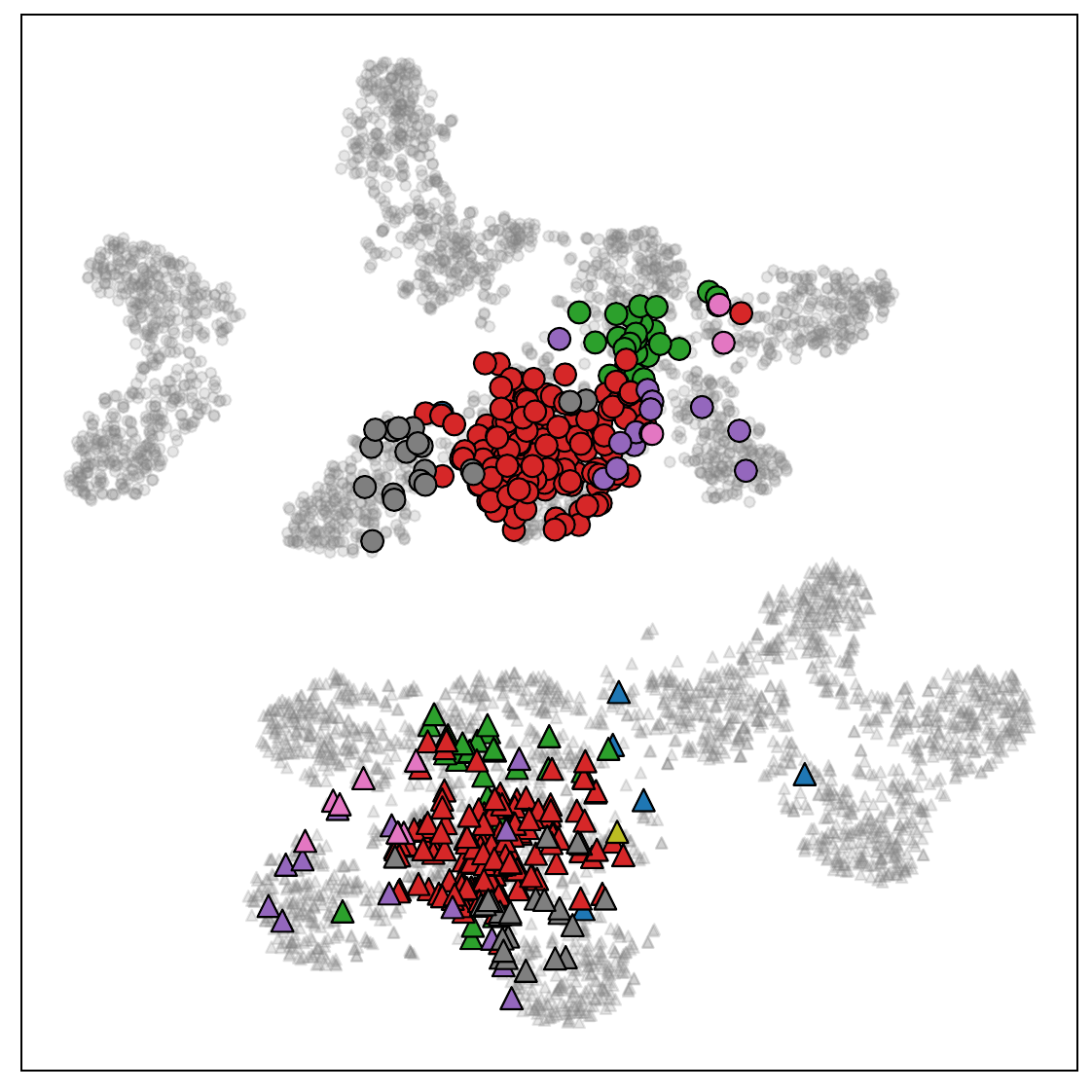}
        \end{minipage}
        \label{fig:predVis_PostF_IS}
    }

\caption{Figure~\ref{fig:predsDistCifar10_uncls_5} shows the unlearned model's predictive label distribution for $\mathcal{D}_f$ of the test set when the 5-th class in the CIFAR-10 (``dog'')  is forgetting class. The blue bar represents the predictive label distribution of the \textit{Retrain} model for the samples in $D_f$. The y-axis is the ratio. The $D_{KL}$ is the predictive distribution disparity between the Retrain model and each method. Figure~\ref{fig:predsDistCifar10_uncls_5}-\ref{fig:predVis_PostF_IS} show representation of test data. Circles are output by the \textit{Retrain} and triangles are output by each method. The grey background shows samples in $\mathcal{D}_r$ and colored points present predicted labels of samples in $\mathcal{D}_f$.}
\label{fig:unlearnedModelRes}
\end{figure*}

\section{More Observation on Unlearned Models}
\label{sec:unlearnedModelRes}

We further observed the output of the unlearned model obtained by each method on the forgetting samples in the test dataset. Specifically, we initially examined the discrepancy between the predictive distribution $P_{un}$ of the unlearned model, obtained by each method for the samples in the $\mathcal{D}_f$, and the predictive distribution $P_{re}$ of the retrained model for these samples. We then assessed the divergence between $P_{re}$ and $P_{un}$ by calculating their KL divergence, where a smaller KL divergence indicates that the unlearned model is more analogous to the retrained model in terms of the distribution of output labels. The ``dog'', also the 5th class in CIFAR-10, was selected as a case study because, as illustrated in Table~\ref{tab:detailRes}, GKT exhibits its optimal performance under this setting. As shown in Figure~\ref{fig:predsDistCifar10_uncls_5}, the proposed methods exhibit markedly reduced $D_{KL}$, except \postfilter, which displays a $D_{KL}$ value comparable to that of GKT. This observation suggests that the unlearned models obtained through our proposed methods exhibit behavior that is more akin to that of the retrained model. Plus, in the second row of Figure~\ref{fig:unlearnedModelRes}, when the ``dog'' is unlearned, the retrained model predicts numerous samples of ``dog'' as ``cat'', ``horse'', and ``deer'', and the representation of ``dog'' samples are also dispersed among these semantic or feature similar classes. In comparison to the retrained model, while the distribution of representations learned by other methods exhibits some divergence from that of the retrained model, the relative relationships between representations remain. Moreover, the relative relationships between representations of different classes learned by our proposed method are more evident.

\begin{table}[th]
\centering
\setlength\tabcolsep{1.5pt}
\begin{tabular}{c|cc|cc}
\toprule
\multirow{2}{*}{Method} & \multicolumn{2}{c|}{AllCNN} & \multicolumn{2}{c}{ResNet18} \\
                        & $A_f$           & $A_r$          & $A_f$            & $A_r$           \\
\midrule
GKT                     & 0.0$\pm$0.0  & 20.49$\pm$0.44             & 0.0$\pm$0.0 & 25.49$\pm$5.25            \\
\inhibsynShort                     & 0.0$\pm$0.0  & \textbf{67.33$\pm$2.93}    & 0.0$\pm$0.0 & \underline{79.86$\pm$4.93}            \\
\postfilterShort              & 0.0$\pm$0.0  & 34.06$\pm$0.49             & 0.0$\pm$0.0 & 45.02$\pm$3.89            \\
\methodname           & 0.0$\pm$0.0  & \underline{66.23$\pm$5.34} & 0.0$\pm$0.0 & \textbf{85.67$\pm$0.65}            \\
\bottomrule
\end{tabular}
\caption{Accuracy results on SVHN when using ZSKT.}
\label{tab:zsktResSVHN}
\end{table}

\begin{figure*}[ht]
\centering
    \subfigure[AllCNN]{
        \begin{minipage}[t]{\linewidth}
            \centering
            \includegraphics[width=\linewidth]{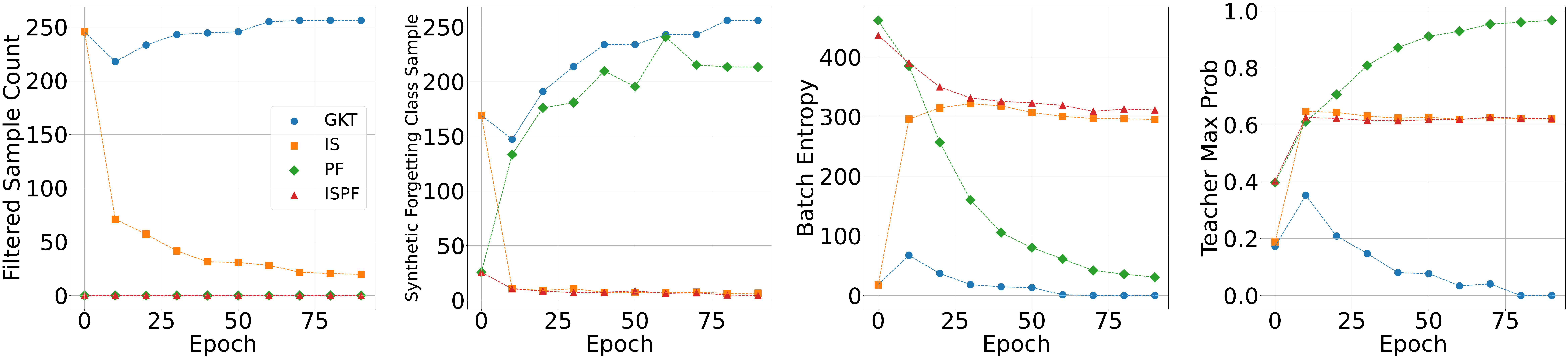}
        \end{minipage}
        \label{fig:zsktSVHNallcnn}
    }

    \subfigure[ResNet18]{
        \begin{minipage}[t]{\linewidth}
            \centering
            \includegraphics[width=\linewidth]{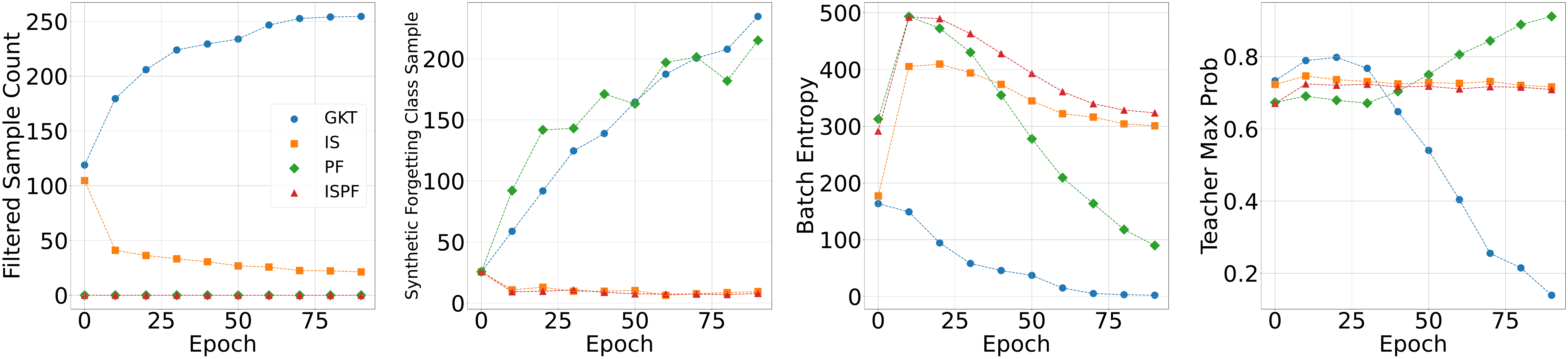}
        \end{minipage}
        \label{fig:zsktSVHNresnet18}
    }

\caption{Results on the SVHN dataset using ZSKT as the fundamental DFKD.}
\label{fig:zsktEffectResSVHN}
\end{figure*}

\section{Replace the Fundamental DFKD Method}
\label{sec:zsktRes}
To the best of our knowledge, ZSKT~\cite{ZSKD} was the first to attempt such a data-free knowledge distillation method accompanied by an adversarial training generator during the distillation process. DFAD~\cite{DFAD} is a similar work from the same period and has validated the effectiveness of such methods on more downstream tasks (e.g., image segmentation).

In contrast to the ZSKT, the more advanced DFQ~\cite{DFQ} introduces an a priori goal for generator learning. This entails utilizing the information recorded in the BatchNorm layer in the teacher to bring the distribution of synthetic samples closer to the distribution of real samples. Additionally, a balancing goal is introduced, which encourages the generator to synthesize a balanced number of samples across classes. In this section, we replace the fundamental DFKD method from DFQ to ZSKT and conduct experiments on the SVHN dataset.

Table~\ref{tab:zsktResSVHN} shows that our proposed methods still outperform the existing method in maintaining the retaining classes' knowledge. As shown in the initial two columns of Figure~\ref{fig:zsktEffectResSVHN}, the number of samples belonging to the forgetting class, as synthesized by GKT, increases rapidly with the training going until all synthesized samples are of the forgetting class. Concurrently, the number of samples filtered out also rises rapidly. This results in a rapid decline in the number of samples utilized for distillation in GKT, as well as the amount of information in each batch (third column of Figure~\ref{fig:zsktEffectResSVHN}). 

We notice that \postfilterShort~without \inhibsynShort~didn't achieve a very good $A_r$ performance. The \postfilterShort~attempts to utilize the retaining-related information in all synthetic samples, including those belonging to the forgetting class. By combining the second and fourth columns of Figure~\ref{fig:zsktEffectResSVHN}, it can be seen that when ZSKT is used as the fundamental DFKD, the \postfilterShort~without \inhibsynShort~synthesizes an increasing number of samples belonging to the forgetting class, and that these samples have a high degree of confidence. We believe that this is because the training of ZSKT's generator has no balancing objective, resulting in the generator having no opportunity to explore the distribution of other classes. Furthermore, the third column demonstrates that the amount of information within each batch of \postfilterShort~without \inhibsynShort~also decreases rapidly as training proceeds, thereby reducing distillation efficiency.

\begin{table}[th]
\centering
\setlength\tabcolsep{2.5pt}
\begin{tabular}{c|c|cc|cc}
\toprule
\multirow{2}{*}{Arch}     & \multirow{2}{*}{Method} & \multicolumn{2}{c}{SVHN} & \multicolumn{2}{c}{CIFAR-10} \\
                          &                         & $A_r$         & MIA$_{II}$        & $A_r$           & MIA$_{II}$          \\
\midrule
\multirow{5}{*}{\rotatebox{90}{AllCNN}}   & Retrain         & 94.14$\pm$0.13                & 33.28             & 91.39$\pm$0.04                & 37.34             \\ \cline{2-6}
                          & PD                              & 86.58$\pm$0.61                & 20.35             & 76.16$\pm$0.21                & \underline{40.55}             \\
                          & PD+\inhibsynShort               & \underline{92.63$\pm$0.08}    & 24.58             & \textbf{86.06$\pm$0.07}       & 41.68             \\
                          & \postfilterShort                & 91.15$\pm$0.85                & \textbf{34.24}    & 76.76$\pm$0.06                & 26.36            \\
                          & \postfilterShort+\inhibsynShort & \textbf{92.68$\pm$0.13}       & \underline{34.54} & \underline{86.02$\pm$0.03}    & \textbf{36.14}            \\
\midrule
\multirow{5}{*}{\rotatebox{90}{ResNet18}}   & Retrain       & 94.43$\pm$0.03          & 36.51           & 92.41$\pm$0.23            & 30.81             \\ \cline{2-6}
                          & PD                              & 90.64$\pm$0.55          & 4.25            & 77.64$\pm$2.16            & \underline{36.27}             \\
                          & PD+\inhibsynShort               & \underline{91.88$\pm$0.21}          & 7.62            & \underline{81.86$\pm$0.23}            & 36.84             \\
                          & \postfilterShort                & 91.61$\pm$0.41          & \underline{32.28}           & 81.21$\pm$0.72            & 22.73             \\
                          & \postfilterShort+\inhibsynShort & \textbf{91.92$\pm$0.23}          & \textbf{33.67}           & \textbf{83.33$\pm$0.81}            & \textbf{29.36}             \\
\bottomrule
\end{tabular}
\caption{Results when replacing \postfilter.}
\label{tab:placePostFwithDF}
\end{table}

\section{Replace the \postfilter}
\label{sec:replacePostF}

\postfilter~prevents the forgetting-related information by modifying the output of the teacher. A simpler implementation is setting the value of $T(\widetilde{x})$ corresponding to the forgetting class as 0 and renormalizing the rest, which we refer to as Probability Distribution (PD). In this section, we compare the PD with our proposed implementation of \postfilter.

As presented in Table~\ref{tab:placePostFwithDF}, our proposed implementation of \postfilter~demonstrates superior performance compared to PD's $A_r$. With regard to the unlearning, compared with the PD, our proposed \postfilter~performs much closer to the retrained model. \chenhao{The underlying reason is likely to be related to the nature of knowledge distillation, where the real-valued outputs of the teacher model are essential. Directly setting all forgetting classes to zero after the softmax step can lead to hard clipping of the teacher output and distortion of the supervision signals on the retaining classes provided by the teacher model, leading to decreased retaining performance.}

\section{Synthetic Samples}
\label{sec:synSample}
\begin{figure*}[th]
\centering
    \subfigure[GKT]{
        \begin{minipage}[t]{0.23\linewidth}
            \centering
            \includegraphics[width=\linewidth]{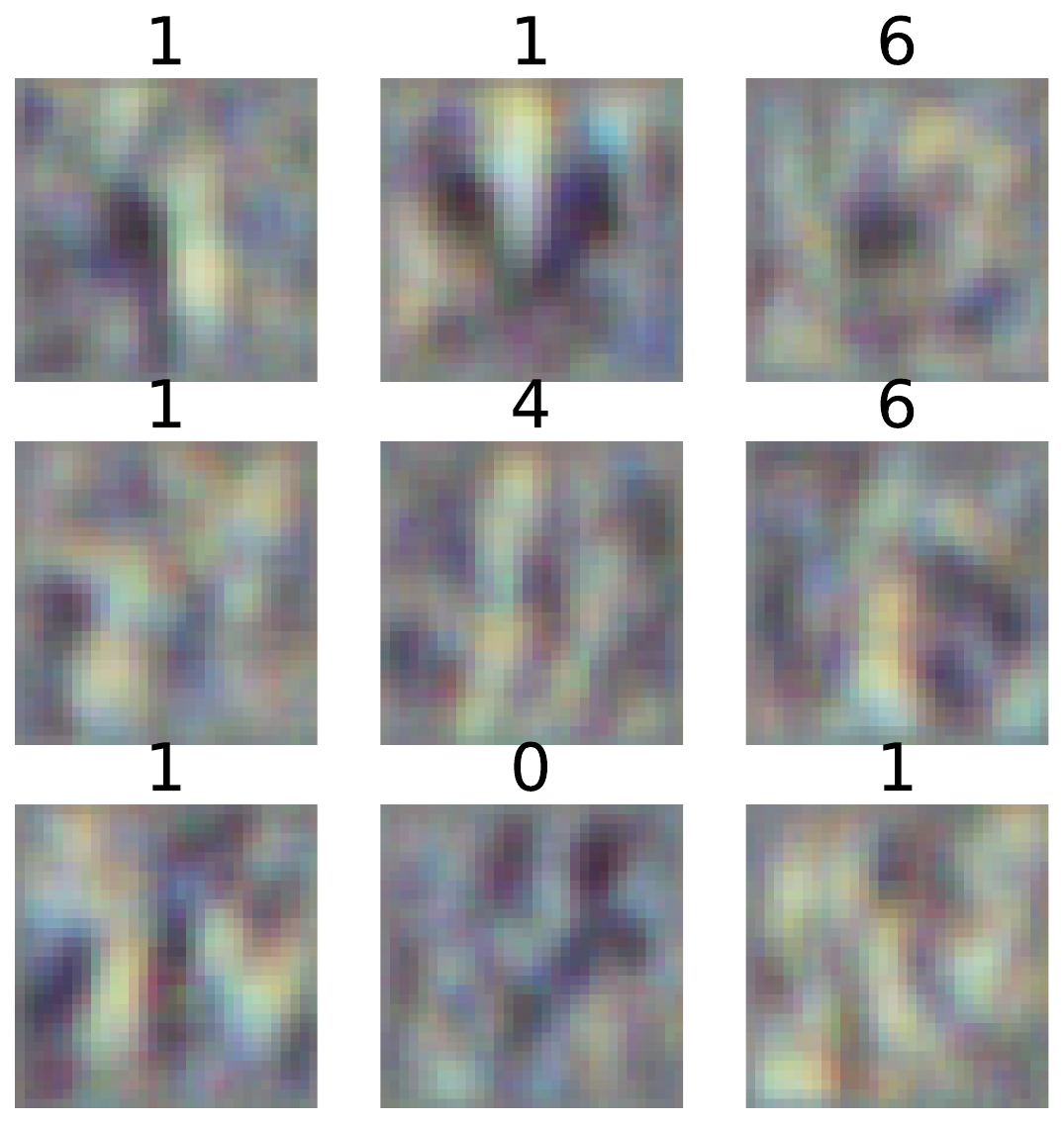}
        \end{minipage}
        \label{fig:sampleVisGKT}
    }
    \hfill
    \subfigure[\inhibsynShort]{
        \begin{minipage}[t]{0.23\linewidth}
            \centering
            \includegraphics[width=\linewidth]{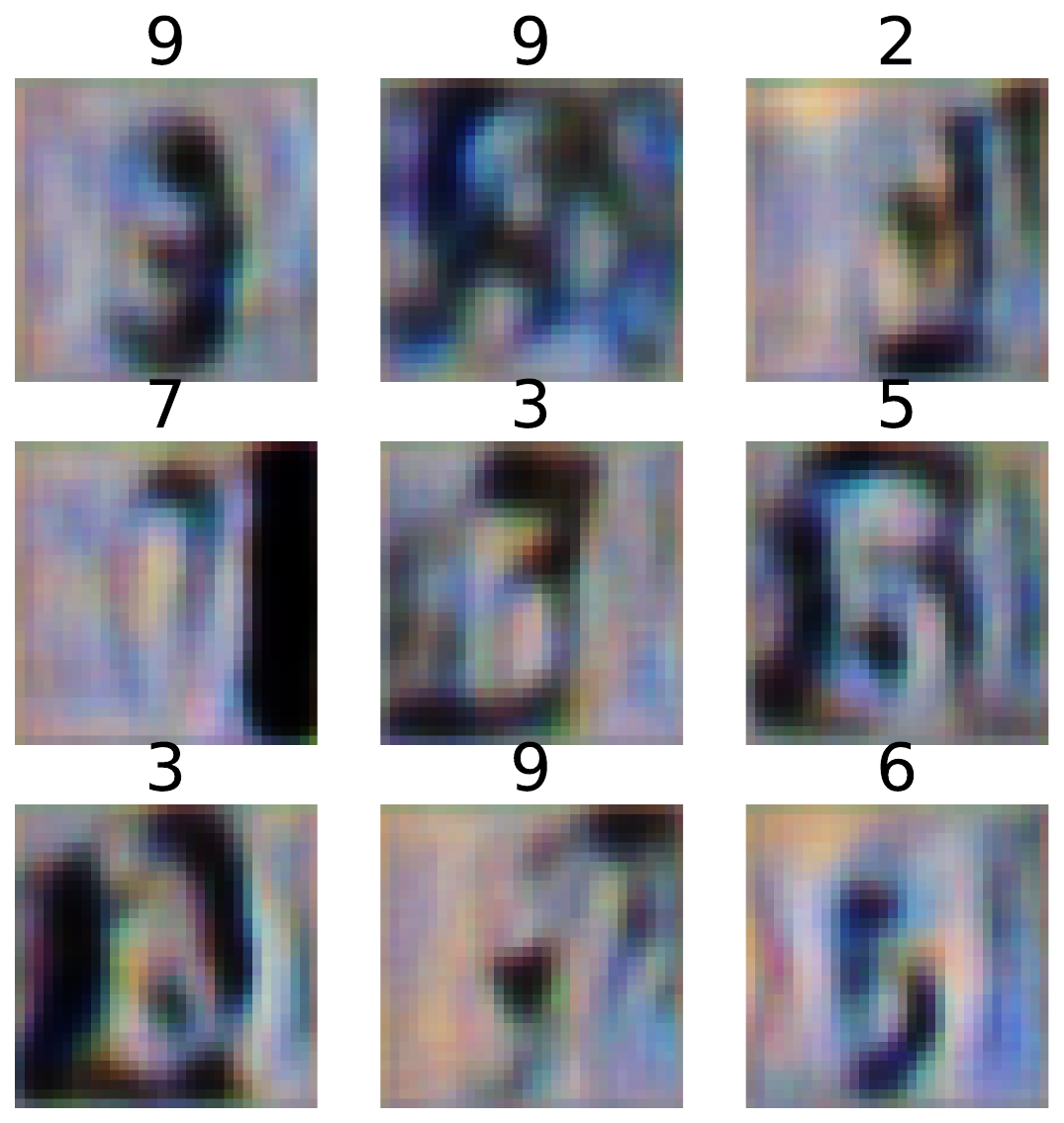}
        \end{minipage}
        \label{fig:sampleVisIS}
    }
    \hfill
    \subfigure[\postfilterShort]{
        \begin{minipage}[t]{0.23\linewidth}
            \centering
            \includegraphics[width=\linewidth]{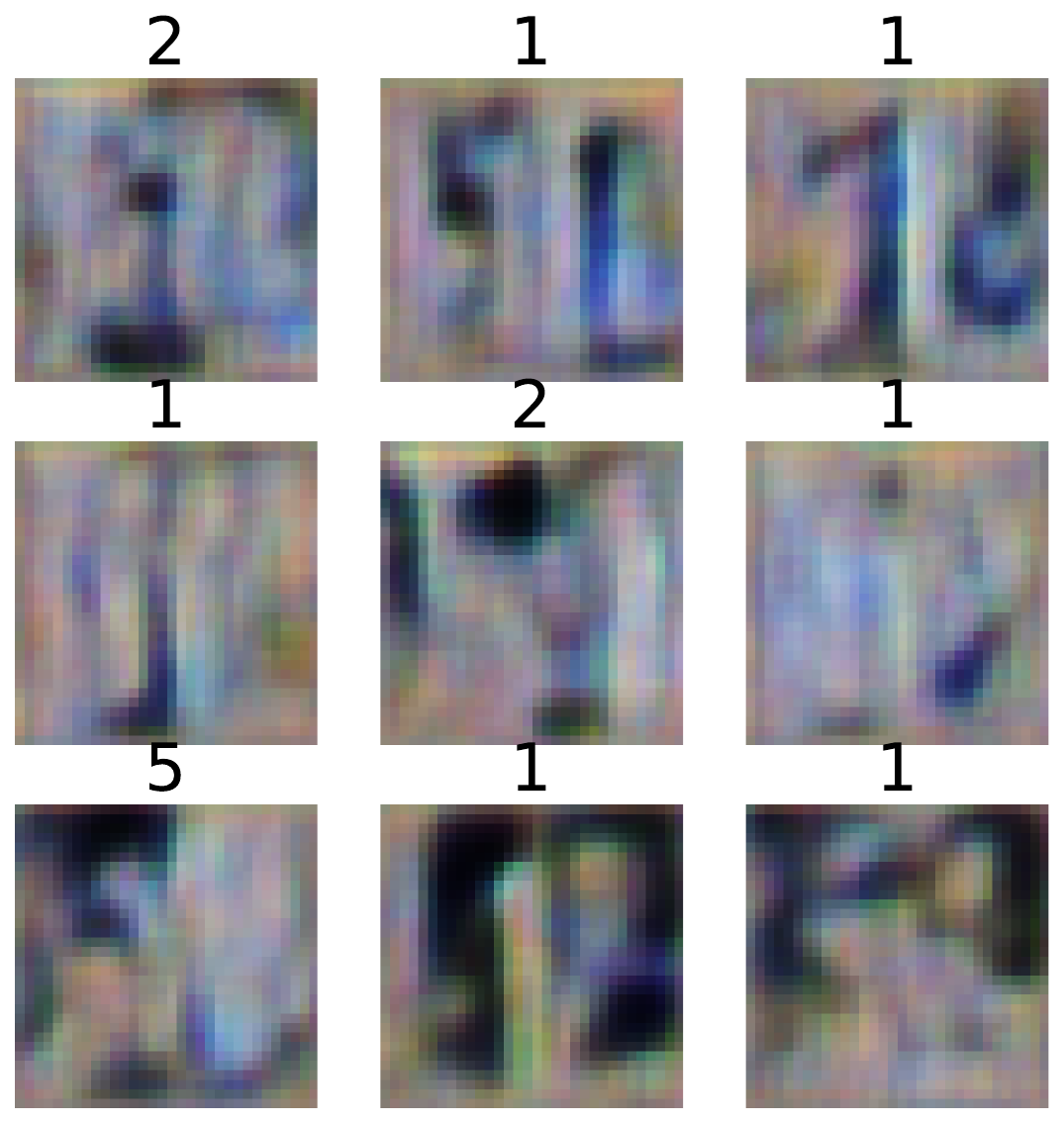}
        \end{minipage}
        \label{fig:sampleVisPostF}
    }
    \hfill
    \subfigure[\methodname]{
        \begin{minipage}[t]{0.23\linewidth}
            \centering
            \includegraphics[width=\linewidth]{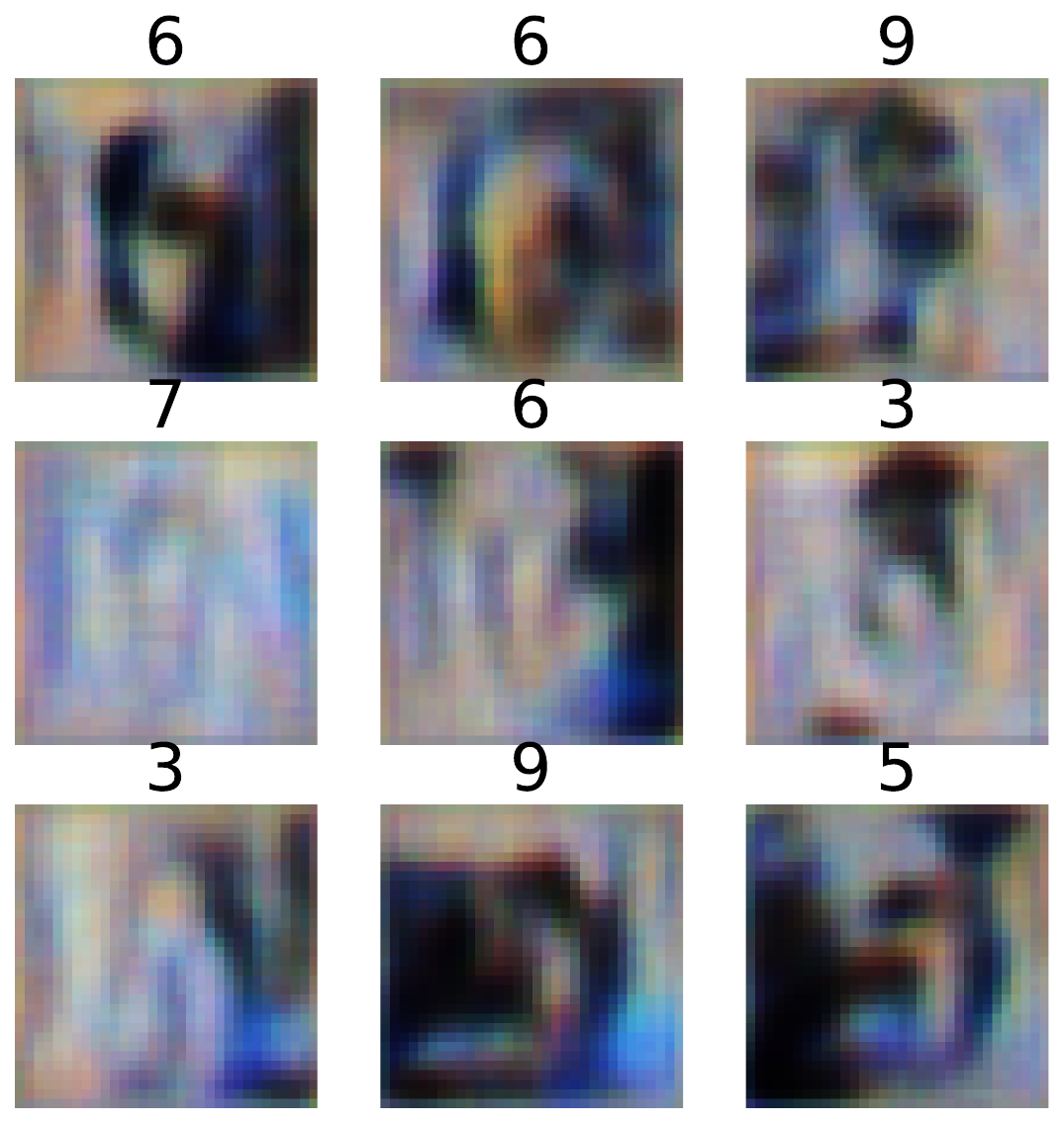}
        \end{minipage}
        \label{fig:sampleVisPostFIS}
    }
\caption{Synthetic samples visualization. Unlearning ``1'' of SVHN with AllCNN as the network. \chenhao{It is demonstrated that IS contributes to generating less data for forgetting classes.}}
\label{fig:synSamples}
\end{figure*}

\begin{figure*}[th]
\centering
    \subfigure[GKT]{
        \begin{minipage}[t]{0.23\linewidth}
            \centering
            \includegraphics[width=\linewidth]{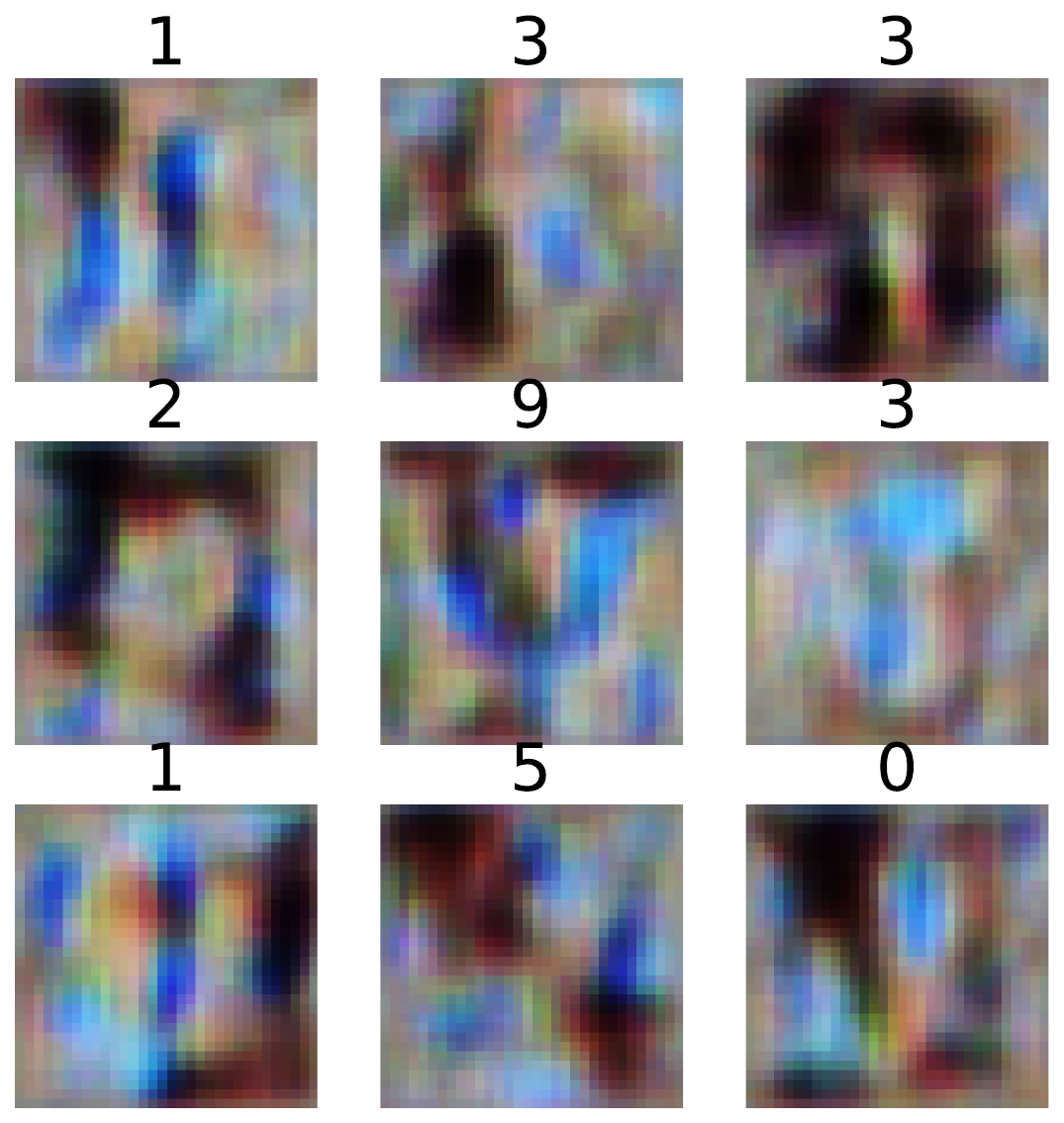}
        \end{minipage}
        \label{fig:sampleVisGKTRes}
    }
    \hfill
    \subfigure[\inhibsynShort]{
        \begin{minipage}[t]{0.23\linewidth}
            \centering
            \includegraphics[width=\linewidth]{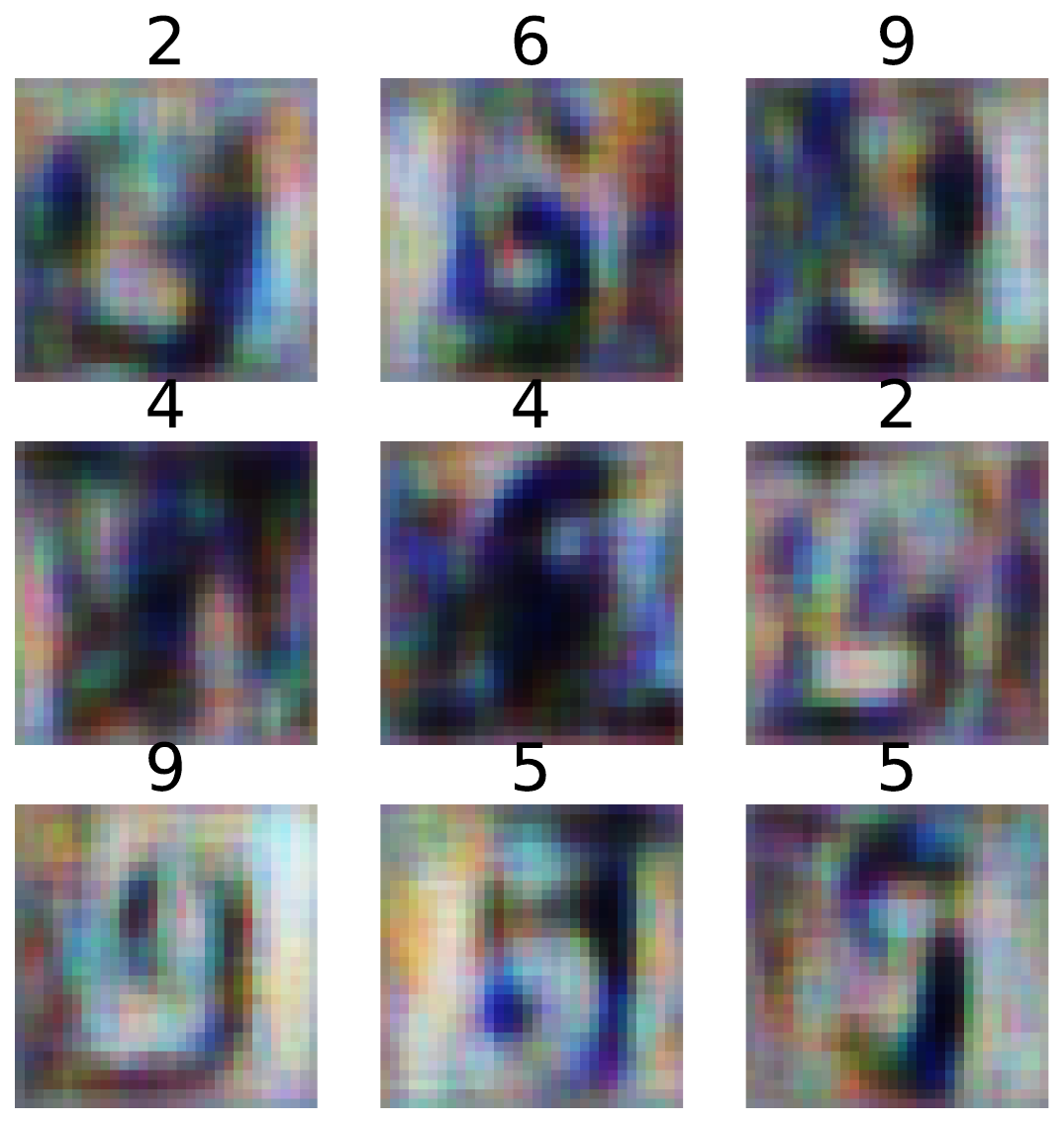}
        \end{minipage}
        \label{fig:sampleVisISRes}
    }
    \hfill
    \subfigure[\postfilterShort]{
        \begin{minipage}[t]{0.23\linewidth}
            \centering
            \includegraphics[width=\linewidth]{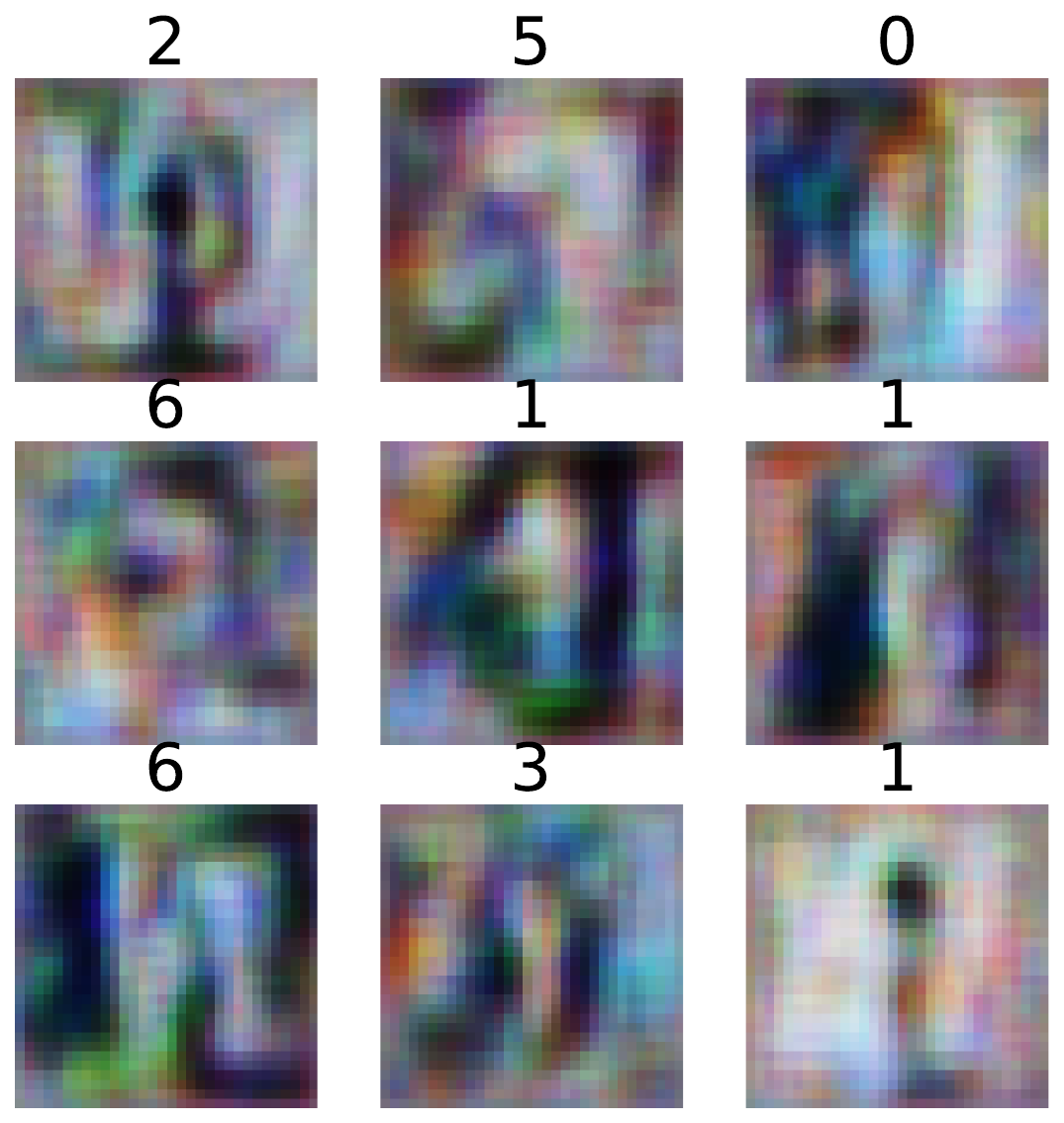}
        \end{minipage}
        \label{fig:sampleVisPostFRes}
    }
    \hfill
    \subfigure[\methodname]{
        \begin{minipage}[t]{0.23\linewidth}
            \centering
            \includegraphics[width=\linewidth]{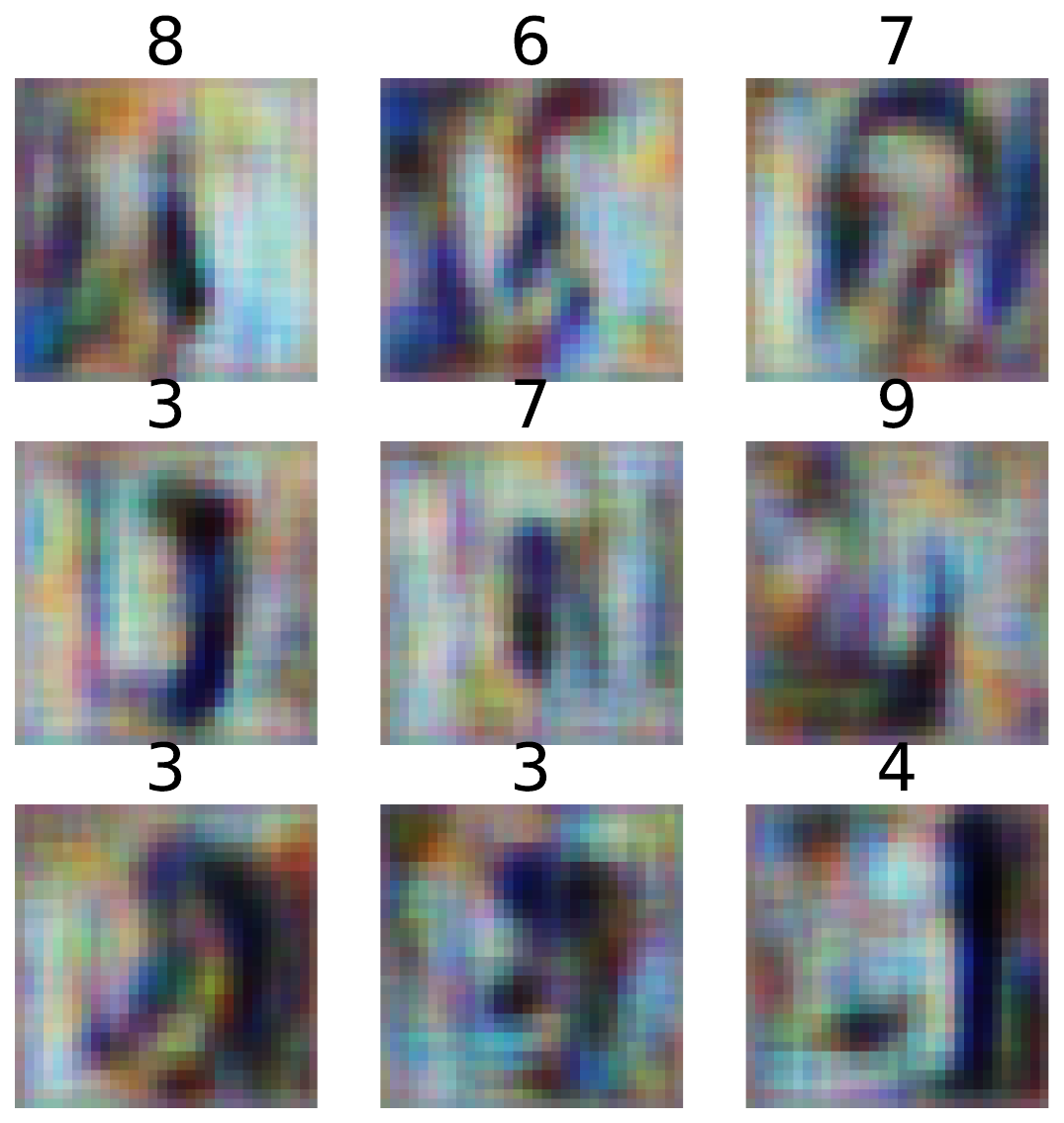}
        \end{minipage}
        \label{fig:sampleVisPostFISRes}
    }
\caption{\chenhao{Synthetic samples visualization. Unlearning ``1'' of SVHN with ResNet18 as the network. It is demonstrated that IS contributes to generating less data for forgetting classes.}}
\label{fig:synSamplesRes}
\end{figure*}

Figure~\ref{fig:synSamples} and Figure~\ref{fig:synSamplesRes} demonstrate samples synthesized by the generators trained using each method. We use the unlearning of ``1'' in SVHM as an example. From these two figures, GKT and \postfilterShort~without \inhibsynShort~produce a considerable number of samples belonging to the forgetting class. The student in GKT is not effectively trained, which hinders the generator from being effectively trained in an adversarial manner. Conversely, the methods utilizing \inhibsynShort~produce a limited amount of characteristics associated with the forgetting class, and synthetic samples exhibit greater divergence.

\begin{table*}[t]
\centering
\setlength\tabcolsep{2.5pt}
\begin{tabular}{c|c|cccccccccc}
\toprule
\multirow{2}{*}{D-A} & \multirow{2}{*}{Method}        & \multicolumn{10}{c}{Forgetting Class} \\% & \multirow{2}{*}{AVG} \\ \cline{3-12}
                              &                                & 0  & 1 & 2 & 3 & 4 & 5 & 6 & 7 & 8 & 9 \\%&                      \\
\midrule

\multirow{10}{*}{\rotatebox{90}{SVHN-AllCNN}} 
 & \multirow{2}{*}{ BlockF } & 92.71$_{0.21}$ & 91.52$_{0.03}$ & 91.82$_{0.03}$ & 93.13$_{0.25}$ & 92.08$_{0.25}$ & 92.11$_{0.08}$ & 92.45$_{0.17}$ & 92.45$_{0.47}$ & 92.97$_{0.2}$ & 92.31$_{0.05}$\\
 & &92.46$_{0.6}$ &94.54$_{0.17}$ &96.34$_{0.51}$ &87.65$_{0.42}$ &92.13$_{0.14}$ &93.1$_{1.32}$ &86.49$_{4.0}$ &90.44$_{2.97}$ &79.28$_{1.45}$ &90.85$_{2.13}$\\ \cline{2-12}
 & \multirow{2}{*}{ GKT } & 90.55$_{0.2}$ & 19.78$_{0.04}$ & 23.69$_{0.39}$ & 33.6$_{12.16}$ & 19.39$_{1.72}$ & 69.7$_{9.24}$ & 78.5$_{2.07}$ & 84.49$_{0.59}$ & 89.38$_{0.0}$ & 54.58$_{23.01}$\\
 & &0.0$_{0.0}$ &0.0$_{0.0}$ &0.0$_{0.0}$ &0.0$_{0.0}$ &0.0$_{0.0}$ &0.0$_{0.0}$ &0.0$_{0.0}$ &0.0$_{0.0}$ &0.0$_{0.0}$ &0.0$_{0.0}$\\ \cline{2-12}
 & \multirow{2}{*}{ \inhibsynShort } & 91.74$_{0.21}$ & 89.68$_{1.28}$ & 89.85$_{0.21}$ & 91.36$_{0.29}$ & 91.07$_{0.1}$ & 91.03$_{0.4}$ & 90.54$_{0.45}$ & 91.64$_{0.11}$ & 91.13$_{0.17}$ & 89.76$_{0.59}$\\
 & &0.0$_{0.0}$ &0.0$_{0.0}$ &0.0$_{0.0}$ &0.0$_{0.0}$ &0.0$_{0.0}$ &0.0$_{0.0}$ &0.0$_{0.0}$ &0.0$_{0.0}$ &0.0$_{0.0}$ &0.0$_{0.0}$\\ \cline{2-12}
 & \multirow{2}{*}{ \postfilterShort } & 92.4$_{0.22}$ & 91.38$_{0.03}$ & 90.26$_{1.32}$ & 92.11$_{0.27}$ & 91.72$_{0.4}$ & 90.49$_{0.23}$ & 92.29$_{0.34}$ & 91.36$_{0.82}$ & 92.92$_{0.03}$ & 92.29$_{0.29}$\\
 & &0.0$_{0.0}$ &0.0$_{0.0}$ &0.0$_{0.0}$ &0.0$_{0.0}$ &0.0$_{0.0}$ &0.0$_{0.0}$ &0.0$_{0.0}$ &0.0$_{0.0}$ &0.0$_{0.0}$ &0.0$_{0.0}$\\ \cline{2-12}
 & \multirow{2}{*}{ \methodname } & \textbf{92.58$_{0.22}$} & \textbf{92.77$_{0.04}$} & \textbf{92.15$_{0.26}$} & \textbf{93.42$_{0.12}$} & \textbf{92.63$_{0.16}$} & \textbf{92.6$_{0.1}$} & \textbf{92.92$_{0.0}$} & \textbf{92.64$_{0.09}$} & \textbf{93.06$_{0.01}$} & \textbf{92.77$_{0.01}$}\\
 & &0.0$_{0.0}$ &0.0$_{0.0}$ &0.0$_{0.0}$ &0.0$_{0.0}$ &0.0$_{0.0}$ &0.0$_{0.0}$ &0.0$_{0.0}$ &0.0$_{0.0}$ &0.0$_{0.0}$ &0.0$_{0.0}$\\

\midrule

\multirow{10}{*}{\rotatebox{90}{SVHN-ResNet18}} 
 & \multirow{2}{*}{ BlockF } & 91.15$_{0.57}$ & 92.18$_{0.09}$ & 91.27$_{0.15}$ & 92.61$_{0.4}$ & 91.38$_{0.06}$ & 91.83$_{0.34}$ & 91.8$_{0.33}$ & 92.01$_{0.21}$ & 92.44$_{0.29}$ & 91.98$_{0.01}$\\
 & &87.18$_{1.18}$ &86.36$_{3.38}$ &79.38$_{1.05}$ &60.74$_{4.46}$ &75.86$_{2.7}$ &78.12$_{0.27}$ &73.32$_{2.5}$ &83.63$_{0.22}$ &46.66$_{13.16}$ &78.4$_{0.47}$\\ \cline{2-12}
 & \multirow{2}{*}{ GKT } & 90.79$_{0.27}$ & 76.06$_{12.65}$ & 88.32$_{2.11}$ & 91.44$_{0.14}$ & 87.13$_{0.81}$ & 89.76$_{0.09}$ & 90.76$_{0.14}$ & 90.53$_{0.46}$ & 91.49$_{0.32}$ & 91.2$_{0.13}$\\
 & &0.0$_{0.0}$ &0.0$_{0.0}$ &0.0$_{0.0}$ &0.0$_{0.0}$ &0.0$_{0.0}$ &0.0$_{0.0}$ &0.0$_{0.0}$ &0.0$_{0.0}$ &0.0$_{0.0}$ &0.0$_{0.0}$\\ \cline{2-12}
 & \multirow{2}{*}{ \inhibsynShort } & 90.82$_{0.36}$ & 90.64$_{0.41}$ & 91.35$_{0.21}$ & 89.18$_{1.39}$ & 91.22$_{0.54}$ & 88.58$_{1.37}$ & 91.25$_{0.41}$ & 91.49$_{0.19}$ & 91.59$_{0.85}$ & 91.34$_{0.21}$\\
 & &0.0$_{0.0}$ &0.0$_{0.0}$ &0.0$_{0.0}$ &0.0$_{0.0}$ &0.0$_{0.0}$ &0.0$_{0.0}$ &0.0$_{0.0}$ &0.0$_{0.0}$ &0.0$_{0.0}$ &0.0$_{0.0}$\\ \cline{2-12}
 & \multirow{2}{*}{ \postfilterShort } & 90.3$_{1.08}$ & 91.94$_{0.22}$ & 91.33$_{0.11}$ & \textbf{92.6$_{0.23}$} & 90.57$_{1.08}$ & \textbf{91.71$_{0.35}$} & \textbf{92.0$_{0.48}$} & 91.71$_{0.51}$ & 92.18$_{0.45}$ & \textbf{91.76$_{0.37}$}\\
 & &0.0$_{0.0}$ &0.0$_{0.0}$ &0.0$_{0.0}$ &0.0$_{0.0}$ &0.0$_{0.0}$ &0.0$_{0.0}$ &0.0$_{0.0}$ &0.0$_{0.0}$ &0.0$_{0.0}$ &0.0$_{0.0}$\\ \cline{2-12}
 & \multirow{2}{*}{ \methodname } & \textbf{91.74$_{0.05}$} & \textbf{92.55$_{0.14}$} & \textbf{92.05$_{0.12}$} & 91.58$_{0.66}$ & \textbf{92.02$_{0.28}$} & 91.12$_{0.58}$ & 91.9$_{0.77}$ & \textbf{92.19$_{0.19}$} & \textbf{92.4$_{0.29}$} & 91.6$_{0.22}$\\
 & &0.0$_{0.0}$ &0.0$_{0.0}$ &0.0$_{0.0}$ &0.0$_{0.0}$ &0.0$_{0.0}$ &0.0$_{0.0}$ &0.0$_{0.0}$ &0.0$_{0.0}$ &0.0$_{0.0}$ &0.0$_{0.0}$\\

\midrule

\multirow{10}{*}{\rotatebox{90}{CIFAR10-AllCNN}} 
 & \multirow{2}{*}{ BlockF } & 84.26$_{1.22}$ & 84.32$_{0.52}$ & 86.35$_{0.35}$ & 86.25$_{0.42}$ & 84.37$_{0.34}$ & 86.64$_{0.71}$ & 84.47$_{0.41}$ & 84.58$_{0.32}$ & 84.73$_{0.38}$ & 84.81$_{0.32}$\\
 & &70.2$_{8.1}$ &66.95$_{0.95}$ &46.9$_{7.5}$ &56.65$_{0.65}$ &65.9$_{1.3}$ &49.35$_{3.35}$ &72.1$_{1.0}$ &68.1$_{1.3}$ &79.8$_{7.6}$ &80.75$_{1.85}$\\ \cline{2-12}
 & \multirow{2}{*}{ GKT } & 51.12$_{0.89}$ & 71.11$_{3.7}$ & 49.27$_{3.64}$ & 38.74$_{0.41}$ & 50.22$_{1.53}$ & 73.17$_{0.76}$ & 64.34$_{2.82}$ & 50.91$_{7.74}$ & 63.07$_{0.42}$ & 64.21$_{0.73}$\\
 & &0.0$_{0.0}$ &0.0$_{0.0}$ &0.0$_{0.0}$ &0.0$_{0.0}$ &0.0$_{0.0}$ &0.0$_{0.0}$ &0.0$_{0.0}$ &0.0$_{0.0}$ &0.0$_{0.0}$ &0.0$_{0.0}$\\ \cline{2-12}
 & \multirow{2}{*}{ \inhibsynShort } & 85.02$_{0.24}$ & 84.23$_{0.38}$ & 86.13$_{0.26}$ & 85.05$_{0.34}$ & 84.24$_{0.05}$ & 86.31$_{0.27}$ & 84.74$_{0.28}$ & 84.68$_{0.06}$ & 84.24$_{0.17}$ & 84.66$_{0.19}$\\
 & &0.0$_{0.0}$ &0.0$_{0.0}$ &0.0$_{0.0}$ &0.0$_{0.0}$ &0.0$_{0.0}$ &0.0$_{0.0}$ &0.0$_{0.0}$ &0.0$_{0.0}$ &0.0$_{0.0}$ &0.0$_{0.0}$\\ \cline{2-12}
 & \multirow{2}{*}{ \postfilterShort } & 80.08$_{1.41}$ & 70.47$_{1.38}$ & 81.67$_{1.72}$ & 76.8$_{1.02}$ & 77.34$_{0.69}$ & 72.58$_{1.25}$ & 81.78$_{1.12}$ & 76.75$_{0.99}$ & 75.27$_{1.96}$ & 74.89$_{2.61}$\\
 & &0.0$_{0.0}$ &0.0$_{0.0}$ &0.0$_{0.0}$ &0.0$_{0.0}$ &0.0$_{0.0}$ &0.0$_{0.0}$ &0.0$_{0.0}$ &0.0$_{0.0}$ &0.0$_{0.0}$ &0.0$_{0.0}$\\ \cline{2-12}
 & \multirow{2}{*}{ \methodname } & \textbf{86.01$_{0.02}$} & \textbf{85.02$_{0.33}$} & \textbf{86.82$_{0.41}$} & \textbf{87.88$_{0.25}$} & \textbf{86.1$_{0.13}$} & \textbf{87.21$_{0.45}$} & \textbf{85.44$_{0.38}$} & \textbf{85.39$_{0.1}$} & \textbf{84.78$_{0.22}$} & \textbf{85.52$_{0.12}$}\\
 & &0.0$_{0.0}$ &0.0$_{0.0}$ &0.0$_{0.0}$ &0.0$_{0.0}$ &0.0$_{0.0}$ &0.0$_{0.0}$ &0.0$_{0.0}$ &0.0$_{0.0}$ &0.0$_{0.0}$ &0.0$_{0.0}$\\

\midrule

\multirow{10}{*}{\rotatebox{90}{CIFAR10-ResNet18}} 
 & \multirow{2}{*}{ BlockF } & 82.71$_{0.76}$ & 83.36$_{0.58}$ & 84.27$_{0.16}$ & 80.2$_{2.27}$ & 76.29$_{6.42}$ & 81.58$_{2.26}$ & 82.95$_{0.16}$ & 82.51$_{0.78}$ & 81.45$_{1.84}$ & 79.58$_{2.54}$\\
 & &65.15$_{3.15}$ &47.25$_{4.85}$ &37.1$_{3.9}$ &11.85$_{5.05}$ &40.8$_{8.9}$ &29.7$_{1.0}$ &54.0$_{0.4}$ &52.6$_{6.2}$ &69.65$_{3.15}$ &76.95$_{1.25}$\\ \cline{2-12}
 & \multirow{2}{*}{ GKT } & 53.08$_{5.72}$ & 79.2$_{0.28}$ & 43.0$_{27.42}$ & 43.3$_{29.44}$ & 54.46$_{9.66}$ & 46.58$_{7.54}$ & 80.48$_{1.09}$ & 64.13$_{6.52}$ & 68.54$_{1.45}$ & 48.34$_{2.28}$\\
 & &0.0$_{0.0}$ &0.0$_{0.0}$ &0.0$_{0.0}$ &0.0$_{0.0}$ &0.0$_{0.0}$ &0.0$_{0.0}$ &0.0$_{0.0}$ &0.0$_{0.0}$ &0.0$_{0.0}$ &0.0$_{0.0}$\\ \cline{2-12}
 & \multirow{2}{*}{ \inhibsynShort } & 79.21$_{4.42}$ & 82.57$_{0.27}$ & 81.11$_{0.72}$ & 78.79$_{1.21}$ & \textbf{80.96$_{1.76}$} & 78.94$_{5.13}$ & 79.58$_{4.72}$ & 82.32$_{1.07}$ & 81.47$_{1.63}$ & 80.49$_{3.33}$\\
 & &0.0$_{0.0}$ &0.0$_{0.0}$ &0.0$_{0.0}$ &0.0$_{0.0}$ &0.0$_{0.0}$ &0.0$_{0.0}$ &0.0$_{0.0}$ &0.0$_{0.0}$ &0.0$_{0.0}$ &0.0$_{0.0}$\\ \cline{2-12}
 & \multirow{2}{*}{ \postfilterShort } & 80.56$_{0.31}$ & 82.2$_{0.8}$ & 83.58$_{1.09}$ & 79.93$_{1.18}$ & 80.74$_{0.97}$ & 84.39$_{0.23}$ & 83.28$_{0.52}$ & 81.28$_{1.56}$ & 78.37$_{4.14}$ & 77.73$_{3.56}$\\
 & &0.0$_{0.0}$ &0.0$_{0.0}$ &0.0$_{0.0}$ &0.0$_{0.0}$ &0.0$_{0.0}$ &0.0$_{0.0}$ &0.0$_{0.0}$ &0.0$_{0.0}$ &0.0$_{0.0}$ &0.0$_{0.0}$\\ \cline{2-12}
 & \multirow{2}{*}{ \methodname } & \textbf{80.96$_{3.65}$} & \textbf{84.02$_{0.42}$} & \textbf{84.87$_{1.41}$} & \textbf{85.26$_{2.46}$} & 77.58$_{6.14}$ & \textbf{85.21$_{0.31}$} & \textbf{83.71$_{0.11}$} & \textbf{83.84$_{0.54}$} & \textbf{83.96$_{1.06}$} & \textbf{83.88$_{0.5}$}\\
 & &0.0$_{0.0}$ &0.0$_{0.0}$ &0.0$_{0.0}$ &0.0$_{0.0}$ &0.0$_{0.0}$ &0.0$_{0.0}$ &0.0$_{0.0}$ &0.0$_{0.0}$ &0.0$_{0.0}$ &0.0$_{0.0}$\\

\bottomrule
\end{tabular}
\caption{Detailed unlearned model performance when each class is the forgetting class. The first row of results for each method is the $A_r$ and the second row is the $A_f$ and they are reported as percentages (\%) in the form of MEAN$_{STD}$. The \textbf{bold} record indicates the best result.}
\label{tab:detailRes}
\end{table*}

\end{document}